\documentclass[10pt,twocolumn,letterpaper]{article}

\usepackage[pagenumbers]{cvpr} %

\newcommand{\method}{Motion 3-to-4\xspace}

\definecolor{yellow}{rgb}{1,1, 0.6}
\definecolor{lightyellow}{rgb}{1,1, 0.8}
\definecolor{orange}{rgb}{1, 0.8, 0.6}
\definecolor{red}{rgb}{1, 0.6, 0.6}
\definecolor{darkblue}{rgb}{0.2, 0.2, 0.8}

\definecolor{intBlueL}{RGB}{117,121,182}
\definecolor{intBlueLL}{RGB}{223,224,236}
\definecolor{intGrayL}{RGB}{135,135,135}
\definecolor{intGrayLL}{RGB}{227,227,227}
\definecolor{intCrimL}{RGB}{112, 59, 84}
\definecolor{intCrimLL}{RGB}{222,212,217}
\definecolor{intTealL}{RGB}{89,143,131}
\definecolor{intTealLL}{RGB}{218,229,226}

\usepackage[dvipsnames]{xcolor}

\newcommand{\bc}{\mathbf{c}}

\newcommand{\bF}{\mathbf{F}} %

\newcommand{\bn}{\mathbf{n}}

\newcommand{\bP}{\mathbf{P}}

\newcommand{\bV}{\mathbf{V}}

\newcommand{\bx}{\mathbf{x}}\newcommand{\bX}{\mathbf{X}}

\newcommand{\bZ}{\mathbf{Z}}

\newcommand{\cA}{\mathcal{A}}

\newcommand{\figref}[1]{Figure~\ref{#1}}
\newcommand{\secref}[1]{Section~\ref{#1}}

\newcommand{\tabref}[1]{Table~\ref{#1}}

\makeatletter
\DeclareRobustCommand\onedot{\futurelet\@let@token\@onedot}
\def\@onedot{\ifx\@let@token.\else.\null\fi\xspace}
 
\def\ie{i.e\onedot}

\makeatother

\definecolor{yellow}{rgb}{1,1, 0.6}
\definecolor{lightyellow}{rgb}{1,1, 0.8}
\definecolor{orange}{rgb}{1, 0.8, 0.6}
\definecolor{red}{rgb}{1, 0.6, 0.6}

\definecolor{wincolor}{rgb}{0.85, 0.0, 0.0}

\definecolor{darkyellow}{rgb}{0.8, 0.8, 0.5}
\definecolor{darkred}{rgb}{0.7, 0.3, 0.3}
\definecolor{darkgreen}{rgb}{0.3, 0.7, 0.3}
\definecolor{blue}{rgb}{0, 0, 1.0}
\definecolor{green}{rgb}{0, 1.0, 0}
\definecolor{pink}{rgb}{1, 0.4, 0.7}

\newcommand{\duster}{\mbox{DUSt3R}\xspace}

\newcommand{\cuter}{\mbox{CUT3R}\xspace}

\newcommand{\boldparagraph}[1]{\vspace{0.1cm}\noindent{\bf #1}}

\definecolor{customlightgray}{rgb}{0.95, 0.95, 0.95} %
\definecolor{darkgreen}{rgb}{0.0, 0.65, 0.0}
\definecolor{darkred}{rgb}{0.75, 0.0, 0.0} %
\definecolor{darkyellow}{rgb}{0.9, 0.72, 0} %
\definecolor{lightyellow}{rgb}{1, 1, 0.8}

\definecolor{DeltaColor}{rgb}{0.039,0.73,0.71}
\definecolor{SigmaColor}{rgb}{0.98,0.45,0.0}
\definecolor{AlphaColor}{rgb}{0,0,0.8}
\definecolor{BetaColor}{rgb}{0.8,0,0.8}
\definecolor{GammaColor}{rgb}{0.514,0.34,0.224}
\definecolor{EpsilonColor}{rgb}{0.353,0.725,0.906}
\definecolor{PurpleColor}{HTML}{9839ff}
\definecolor{RedColor}{rgb}{0.949,0.275, 0.224}
\definecolor{citecolor}{HTML}{0071bc}

\usepackage{pifont}
\usepackage{bbding}

\definecolor{PurpleColor}{HTML}{8B008B}
\definecolor{OrangeColor}{rgb}{0.914,0.541,0.0.141}
\definecolor{GreenColor}{rgb}{0.137,0.573,0.565}
\definecolor{DeltaColor}{rgb}{0.039,0.73,0.71}
\definecolor{SigmaColor}{rgb}{0.98,0.45,0.0}
\definecolor{AlphaColor}{rgb}{0,0,0.8}
\definecolor{BetaColor}{rgb}{0.8,0,0.8}
\definecolor{GammaColor}{rgb}{0.592,0.851,0.149}
\definecolor{EpsilonColor}{rgb}{0.353,0.725,0.906}
\definecolor{PurpleColor}{HTML}{9839ff}
\definecolor{RedColor}{rgb}{0.949,0.275, 0.224}
\definecolor{citecolor}{HTML}{0071bc}

\newcommand{\cmark}{\textcolor{GreenColor}{\ding{51}}\xspace}
\newcommand{\xmark}{\textcolor{RedColor}{\ding{55}}\xspace}

\newlength\savewidth\newcommand\shline{\noalign{\global\savewidth\arrayrulewidth
  \global\arrayrulewidth 1pt}\hline\noalign{\global\arrayrulewidth\savewidth}}

\definecolor{customcolor3D}{HTML}{FED273}
\definecolor{customcolor2D}{HTML}{BBC990}

\definecolor{cvprblue}{rgb}{0.21,0.49,0.74}
\usepackage[pagebackref,breaklinks,colorlinks,allcolors=cvprblue]{hyperref}

\usepackage{multirow}
\usepackage{pifont}
\usepackage{tabularx}
\usepackage{tcolorbox}
\usepackage{tikz}

\def\paperID{***} %
\def\confName{CVPR}
\def\confYear{2026}

\title{\method: 3D Motion Reconstruction for 4D Synthesis}

\author{
Hongyuan Chen$^{1}$ \quad Xingyu Chen$^{1}$ \quad Youjia Zhang$^{1,2}$ \quad Zexiang Xu$^{3}$ \quad Anpei Chen$^{1}$ \vspace{8pt}\\
$^1$Westlake University \quad $^2$HUST \quad $^3$Hillbot
}
\begin{document}
\twocolumn[{%
\renewcommand\twocolumn[1][]{#1}%
\maketitle
\vspace{-10mm}
\begin{center}
    \centering
    \captionsetup{type=figure}
    \includegraphics[width=1.0\linewidth]{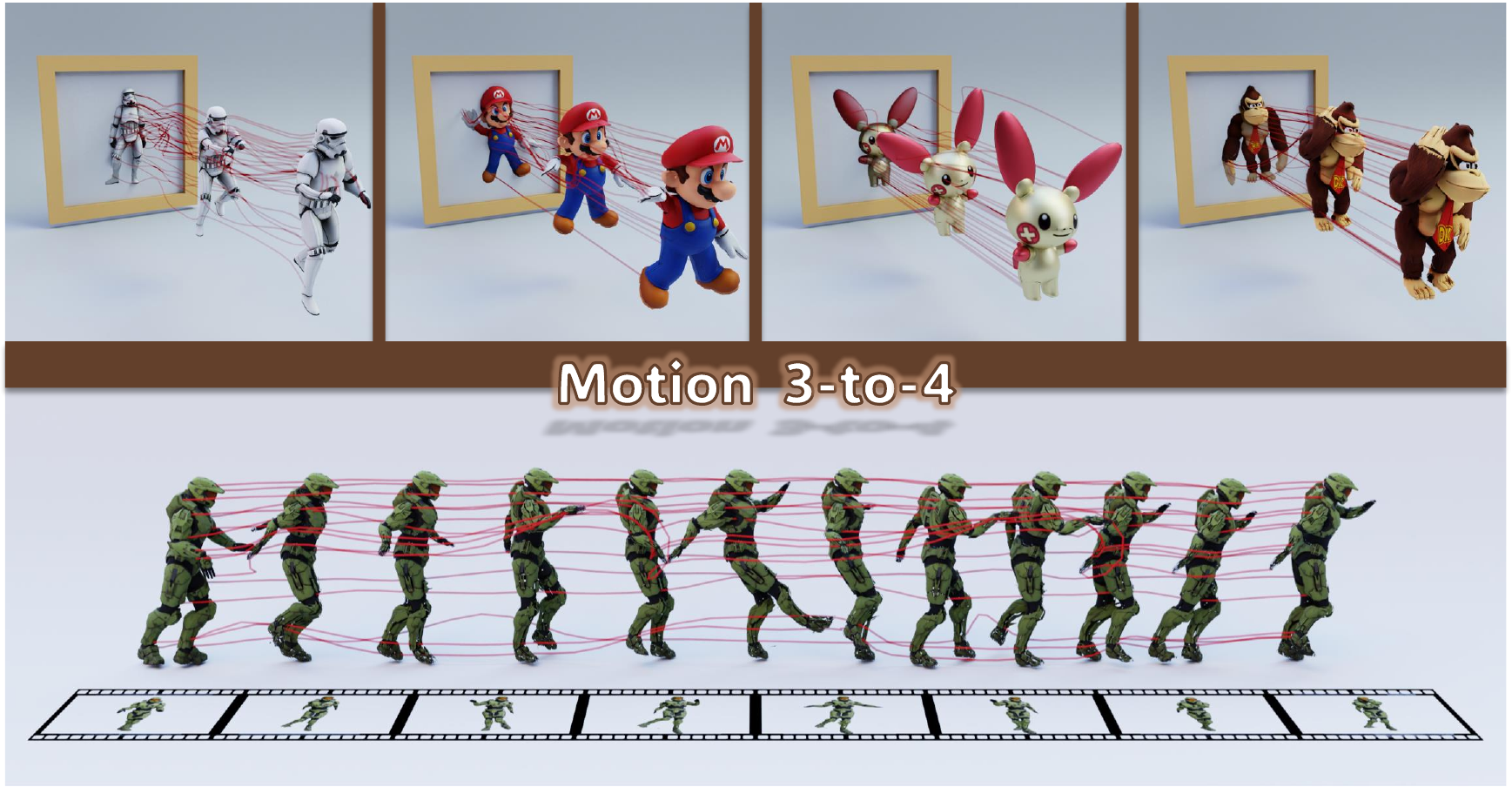}
    \captionof{figure}{\textbf{From a single glance, \method unfolds: weaving time, shape, and movement into living 4D reality.}
    }
    \label{fig:teaser}
\end{center}%
}]

\begin{abstract}
We present Motion 3-to-4, a feed-forward framework for synthesising high-quality 4D dynamic objects from a single monocular video and an optional 3D reference mesh. While recent advances have significantly improved 2D, video, and 3D content generation, 4D synthesis remains difficult due to limited training data and the inherent ambiguity of recovering geometry and motion from a monocular viewpoint.
\method addresses these challenges by decomposing 4D synthesis into static 3D shape generation and motion reconstruction. Using a canonical reference mesh, our model learns a compact motion latent representation and predicts per-frame vertex trajectories to recover complete, temporally coherent geometry. A scalable frame-wise transformer further enables robustness to varying sequence lengths.
Evaluations on both standard benchmarks and a new dataset with accurate ground-truth geometry show that \method  delivers superior fidelity and spatial consistency compared to prior work. Project page is available at \href{https://motion3-to-4.github.io/}{\texttt{\small https://motion3-to-4.github.io/}}.
\end{abstract}
    
\vspace{-2mm}
\section{Introduction}
\label{sec:intro}
The creation of high-fidelity 4D assets--which comprehensively capture both the static shape and the dynamic motion of an object over time--is a critical and highly sought-after capability in fields like virtual reality, cinematography, robotics, and simulation.
Recent advancements in 2D image, video and 3D content synthesis have revolutionized the computer graphics and computer vision communities, achieving high fidelity through scalable datasets and learning frameworks. 
However, 4D reconstruction and generation remain significantly more challenging due to the larger solution space and the inherent complexity of modeling spatial-temporal evolution.

To address these challenges, recent research has advanced along several directions.
One line of work~\cite{zheng2024unified,Consistent4d,4d-fy,Diffusion4d,4diffusion,Sv4d,Cat4d} first generates multi-view videos from text or single-image inputs and then reconstructs 4D assets using dynamic NeRF~\cite{nerf} or Gaussian Splatting~\cite{3dgs}.
While conceptually straightforward, these pipelines rely on lengthy per-instance optimization and are fundamentally constrained by the view inconsistencies inherited from 2D generative models.
Another line of work adopts pretrained 3D generative priors as a foundation for 4D synthesis. For example, V2M4~\cite{V2M4} produces per-frame meshes followed by iterative mesh alignment to recover coherent temporal structure. However, similar to multi-view reconstruction methods, the optimization remains slow and susceptible to temporal artifacts.
Alternatively, several approaches such as GVFD~\cite{gvfd} and AnimateAnyMesh~\cite{AnimateAnyMesh} build motion latent spaces via VAEs and apply the predicted motion to an initial geometry in a feed-forward manner. 
These methods are efficient and can perform inference in seconds. However, VAE-based modeling typically requires large-scale and diverse training data to form a well-structured latent distribution. When trained solely on the limited and narrow 4D datasets, these models struggle to capture complex motion patterns and exhibit poor generalization.
Given the scarcity of high-quality 4D training data, our central idea is to reformulate 4D generation as a combination of 3D shape generation and motion reconstruction. 

In this work, we introduce \method , a feed-forward framework that synthesizes 4D dynamic objects from a single monocular video and an optional reference mesh.
To efficiently tackle this inherently ill-posed problem, we decompose 4D generation into two more tractable components: static 3D shape encoding and dynamic motion reconstruction. Our key insight is to leverage a static mesh, either provided or generated, as a stable reference geometry, and to estimate per-frame 3D motion flow relative to this canonical state.
Given a monocular video and an optional initial mesh as input, \method first performs motion latent learning, jointly encoding the static object shape and the video context into a compact motion representation. Based on this representation, a motion decoder predicts vertex trajectories for queries sampled from the reference mesh, enabling accurate recovery of complete geometry and temporally coherent motion throughout the entire sequence.

We highlight our contributions as follows:
\begin{itemize}
\item \textit{A feed-forward 4D synthesis framework.} We propose a novel feed-forward pipeline that reformulates 4D object synthesis as a motion reconstruction using only monocular video guidance, achieving strong generalization despite the scarcity of high-quality 4D training data.

\item \textit{A scalable architecture.} 
We design a frame-wise transformer architecture that is robust to input meshes of varying resolution and supports flexible processing of videos of arbitrary length.

\item  \textit{Benchmarks.} \looseness=-1 Existing motion reconstruction and 4D generation benchmarks typically provide only multi-view renderings without accurate 3D geometry, making rendering alignment and motion evaluation difficult. To address this limitation, we introduce a new motion-80 benchmark with ground-truth motion, realistic renderings, and geometry. We demonstrate the effectiveness of \method on both this new dataset and a widely used benchmark~\cite{Consistent4d}.
\end{itemize}

\section{Related work}
\boldparagraph{3D Object Generation.}
Early progress in 3D generation largely relied on GAN-based models~\cite{eg3d, get3d, graf, giraffe, sofgan}, allows for fast object crafting for a specific category. Subsequently, optimization-driven methods distill 2D generative priors, typically guided by CLIP scores~\cite{Dream3d, DreamFields} or Score Distillation Sampling (SDS)~\cite{Dreamfusion, Magic3d, Dreamgaussian, sjc, Prolificdreamer, Gaussiandreamer}, or use multi-view generation followed by reconstruction~\cite{Lgm, Mv-adapter, Mvdream, Instantmesh, Instant3d, gao2024graphdreamer}. While conceptually simple, these methods are often time-consuming and struggle to maintain consistent geometry and appearance across views.
Later methods~\cite{Zero-1-to-3, Syncdreamer, Zero123++, Richdreamer} reduce this issue through multi-view fine-tuning on large 3D datasets~\cite{Objaverse,Objaverse-xl}, though reconstruction is still needed to obtain 3D representation.

With the rapid expansion of high-quality 3D data~\cite{Objaverse,Objaverse-xl}, recent work shifts toward direct 3D generation using diffusion-transformers that output explicit 3D representations. These models either tokenize 3D shapes into diffusable latent sets by encoding unstructured point clouds into unordered latent vectors~\cite{3dshape2vecset, Clay, Step1x-3d, Hunyuan3d_25, Hunyuan3d_2}, or adopt voxel-based structured latents that can be decoded into explicit 3D representations~\cite{TRELLIS, Sparc3D, Direct3d-s2, UniLat3D}.
In particular, Hunyuan3D 2.0~\cite{Hunyuan3d_2} achieves high-quality generation of 3D assets with rich geometry and appearance, forming a strong foundation for our 4D asset creation.

\boldparagraph{4D Reconstruction.}
Structure-from-Motion (SfM)~\cite{agarwal2011building,Schoenberger2016CVPR,snavely2006photo} and Simultaneous Localization and Mapping (SLAM)~\cite{mur2015orb,davison2007monoslam,Droid} have long been the foundation for 3D structure and camera pose estimation. 
Although effective, these approaches often struggle with dynamic sequences, leading to performance degeneracy.
To improve robustness in dynamic scenarios, recent work has demonstrated progress toward general dynamic reconstruction by integrating semantic segmentation~\cite{DS-SLAM,MaskRCNN}, optical flows~\cite{zhao2022particlesfm,RAFT}, geometric constraints~\cite{cvd,CasualSAM,Robust-CVD,MegaSaM,pad3r,4dlrm}, generative priors~\cite{hu2024depthcrafter,shao2024learning}. 
Alternatively, point-map-based approaches~\cite{easi3r,monst3r,cut3r} such as \cuter~\cite{cut3r} extend the feedforward 3D foundation model \duster~\cite{dust3r} to dynamic scenes by fine-tuning with dynamic datasets. This yields feedforward 4D reconstruction and can be further scaled to longer sequences \cite{StreamVGGT,Point3R,TTT3R}, novel-view synthesis \cite{som,Neuralsceneflow,Deformable3dgaussians,Mosca}, and 3D tracking \cite{trace,Spatialtrackerv2,St4rtrack,DG-Mesh}.
Despite recent progress, both reconstruction-based paradigms remain inherently non-generative: they cannot hallucinate geometry for occluded or unseen regions, often resulting in incomplete surfaces or missing structures.

\boldparagraph{4D Generation.}
Similar to 3D object generation, early 4D approaches rely on 2D generative priors from multi-view or video models~\cite{zhao2025advances}, or introduce intermediate rigging structures~\cite{UniRig,Magicarticulate,RigNet,Pinocchio,Anymate, he2025category,riganything, ponymation} with skinning-based animation~\cite{animax_video,Pinocchio,Puppeteer,Magicpony,3dfauna,Motion2Motion,makeani,makepose}.
Score Distillation Sampling (SDS)~\cite{Dreamfusion} is also widely adopted to optimize 4D representations from multi-view or video diffusion models~\cite{zheng2024unified, ling2024align, Dreammesh4d, Animate124, Consistent4d, 4d-fy, akd_articulatedk_diffusion,gaussiansee}, but often introduces artifacts.
More recent works use generated multi-view videos as explicit supervision~\cite{4diffusion, Diffusion4d, yang2024diffusion, Sv4d2, Sv4d, Cat4d, jiang2024animate3d}, yet still require slow per-instance optimization.
L4GM~\cite{L4gm} enables faster feed-forward 4D regression from generated multi-view data, but its performance remains limited by scarce 4D data and view-inconsistent 2D priors.

Current 4D generation methods primarily adopt a two-stage pipeline that extends powerful 3D generative models into the temporal dimension.
Notable, V2M4~\cite{V2M4} generates meshes for each frame via pre-trained 3D generative models~\cite{TRELLIS,Hunyuan3d_2}, followed by a post-processing step to align these meshes to form a 4D mesh. This generate-then-align strategy is slow and prone to topology drift due to independently conditioned frame generations.
To improve temporal consistency, ShapeGen4D~\cite{ShapeGen4D} fine-tunes a 3D generator~\cite{Step1x-3d} on 4D data and alleviates the misalignment issue, but still needs time-consuming post-processing.
Parallel efforts such as GVFD~\cite{gvfd} animate Trellis-based~\cite{TRELLIS} 3D Gaussians via a latent diffusion model that learns global temporal changes, enabling motion generation conditioned on monocular videos. However, since the training depends on rendering supervision of the scarcity of 4D data, it yields weak geometry and 3D structure.

In contrast, we take an orthogonal path by combining the strengths of both 3D generation and reconstruction: we reformulate 4D generation as a combination of 3D shape synthesis and motion reconstruction. By generating the entire object and taking motion synthesis as an alignment problem between surface points and video pixels, our method enables efficient motion representation learning and feed-forward prediction without requiring the post-processing alignment steps.

\vspace{-2mm}
\section{\method}
\begin{figure*}[h]      %
  \centering
  \includegraphics[width=0.9\textwidth]{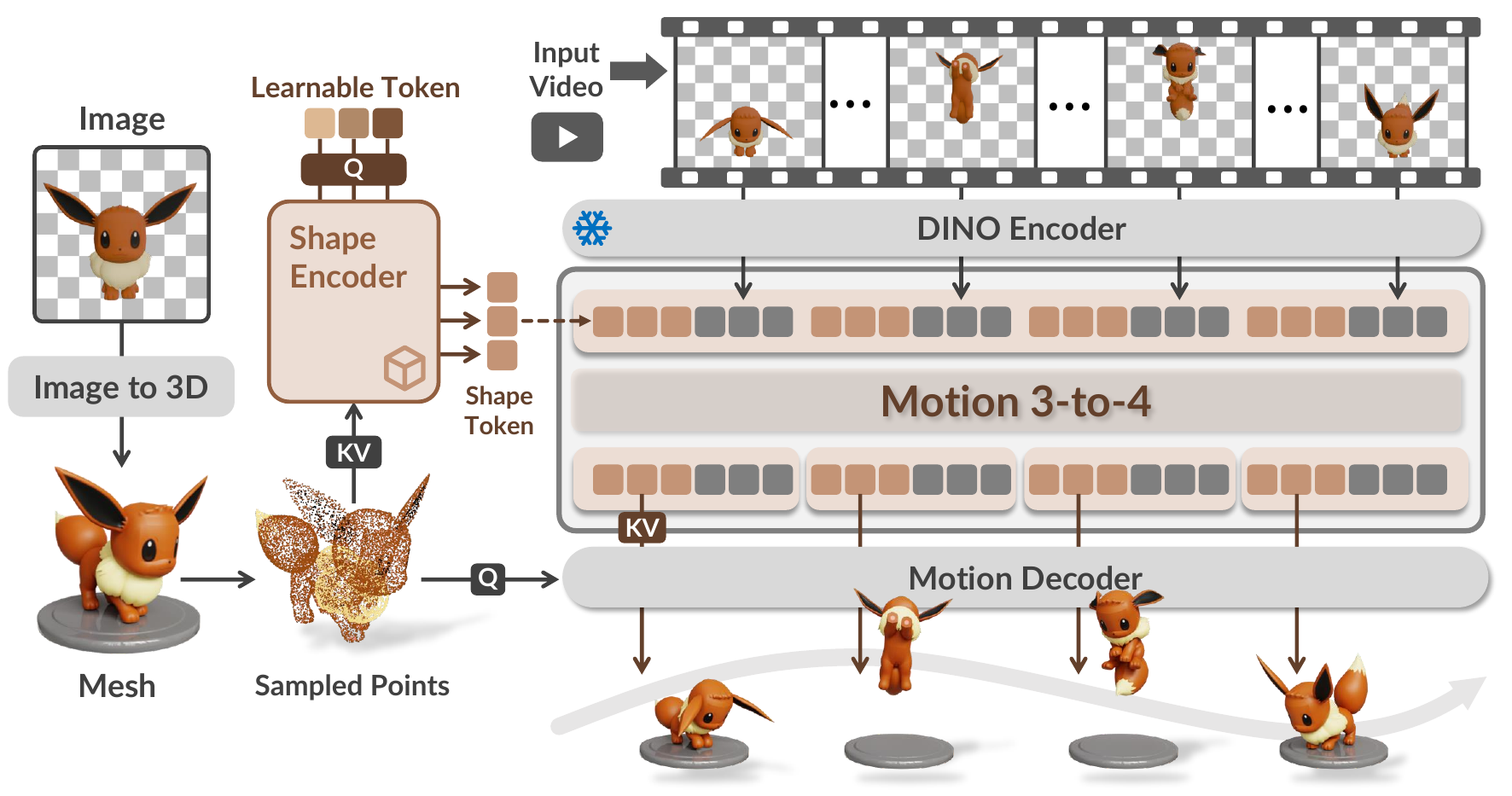}
  \caption{\textbf{An overview of our \method framework for 4D synthesis.} At the core of the framework is a motion–latent learning module consisting of a geometry encoder and a video encoder, which jointly process the input video and sampled points. The resulting latent tokens are decoded into a frame-wise 3D motion flow relative to the first video frame, producing temporally consistent 4D assets.}
  \label{fig:pipeline}
\end{figure*}

In this paper, we aim to efficiently craft 4D assets that encompass complete shape and motion.
Our key idea is to decompose the ill-posed 4D generation problem into static shape generation and dynamic motion reconstruction, enabling the recovery of complete motion flow and geometry, including both visible and unseen surfaces.

To this end, our method takes a monocular video as input and, optionally, an existing mesh asset of the first frame.
If such a mesh is unavailable, we generate one from the initial frame using a pretrained 3D generative model~\cite{Hunyuan3d_2}.
We then estimate per-frame 3D motion flow relative to the first video frame, yielding temporally consistent 4D assets that encapsulate both shape and motion in their entirety, as shown in~\figref{fig:pipeline}.

Our framework consists of two main components: 1) motion latent learning that encodes the static mesh and video frames into a compact representation (\secref{sec:motion_latent_learning}); and 2) motion decoding that regresses per-frame point locations from queries sampled on the static mesh (\secref{sec:motion_decoding}).

\begin{table}[t!]
    \centering
    \setlength\tabcolsep{2pt}
    \scriptsize
    \begin{tabularx}{\columnwidth}{lcccccc}
  \toprule
 {Method} &{Solution} & {FF} &{Motion} &{Mesh} & {Retarget} & {FPS} \\
  \midrule
  Consistent4D~\cite{Consistent4d}  & 
  {MV Gen.} + {3D Rec.} & \xmark & \xmark &  \xmark   & \xmark & 0.1 \\
  SV4D~\cite{Sv4d}  & 
  {MV Gen.} + {3D Rec.} & \xmark & \cmark &  \xmark   & \xmark & 0.1 \\
  L4GM~\cite{L4gm}  & 
  {MV Gen.} + {3D Rec.} & \cmark & \xmark &  \xmark   & \xmark & 7.8 \\
  DreamMesh4D~\cite{Dreammesh4d}  & 
  {MV Gen.} + {3D Rec.} & \xmark & \cmark &  \cmark  & \xmark & 0.1 \\
  V2M4~\cite{V2M4}  & 
  {3D Gen.} + {4D Align} & \xmark & \xmark &  \cmark  & \xmark & 0.1 \\
  ShapeGen4D~\cite{ShapeGen4D}  & 
  {3D Gen.} + {4D Align} & \xmark & \xmark &  \cmark  & \xmark & 0.1 \\
  GVFD~\cite{gvfd}  & 
  {3D Gen.} + {Motion Gen.} & \cmark & \cmark &  \xmark  & \xmark & 0.8 \\
  \midrule
  \textbf{\method{}}  & 
  {3D Gen.} + {Motion Rec.} & \cmark & \cmark &  \cmark  & \cmark & 6.5 \\
  \bottomrule
  \end{tabularx}
  \vspace{-6pt}
  \caption{
   \textbf{An overview of 4D synthesis methods from monocular video.} 
   ``FF'' denotes feed-forward.  ``FPS'' is averaged over 512 frames.
   To address the ill-pose Video-to-4D problem, early pipelines generate multi-view images or videos but suffer from view inconsistency.
   Following frame-wise 3D generative models avoid this issue yet require time-consuming 4D alignment.
   Motion-generation–based methods animate 3D generation, but their generalizability is fundamentally constrained by the limited availability of 4D training data.
   We incorporate static generation and motion reconstruction to learn local surface-to-pixel correspondences for efficient novel shapes and complex motions.
    }
    \label{tab:VFM}
    \vspace{-10pt}
  \end{table}

\vspace{-1mm}

\subsection{Motion Latent Learning}
\label{sec:motion_latent_learning}
In the following, we will introduce a simple yet efficient representation learning framework for 3D motion.

\boldparagraph{Geometric Features.} 
To efficiently capture 3D geometry, we first encode the reference mesh 
$\mathcal{M} \subset \{\mathcal{V}, \mathcal{F}, \mathcal{T} \in \mathbb{R}\}$
with vertices $\mathcal{V}$, faces $\mathcal{F}$, and texture $\mathcal{T}$ into a compact latent representation.
The mesh can either be user-provided or lifted directly from the video using recent image-to-3D generation techniques~\cite{Hunyuan3d_2,TRELLIS}.
To encode shape and appearance, we uniformly sample $N$ surface points $\bX_0 = {(\mathbf{x}_i, \mathbf{n}_i, \mathbf{c}_i)}_{i=1}^{N}$, 
where $\bx_i\in\mathbb{R}^3$ is the 3D coordinate, $\bn_i\in\mathbb{R}^3$ is the surface normal, and $\bc_i\in\mathbb{R}^3$ denotes the RGB color.

Our shape encoder is inspired by 3DShape2VecSet~\cite{3dshape2vecset}, and we compress the sampled points into a compact 1D latent representation by performing cross attention to aggregate shape information. Specifically, we employ a  learnable query set $\cA \in \mathbb{R}^{K \times C}$ of fixed length $K$, where each query token is a $C$-dimensional latent vector. These query tokens act as anchors that attend to and gather information from their neighboring shape samples in $\bX_0$, producing the shape latent representation.
\vspace{-2mm}
\begin{equation}
\bZ_{\bX_0} = \texttt{CrossSelfAttn}(\cA, \text{PointEmb}(\bX_0))
\end{equation}

\noindent where $\text{PointEmb}(\bX_0): \mathbb{R}^9 \rightarrow \mathbb{R}^C$ maps the point label into a high-dimensional positional embedding using an MLP. The aggregated tokens are further refined through a few layers of self-attention transformer blocks to exchange context.

This process embeds the mesh geometry into a low-dimensional latent space, yielding a shape latent $\bZ_{\bX_0} \in \mathbb{R}^{K \times C}$ that retains the essential geometric and semantic structure required for motion reconstruction.

\boldparagraph{Modulation with Video Features.} Next, we aggregate the geometric token $\bZ_{\bX_0}$ with the spatial-temporal video sequence to obtain a motion representation. 

Specifically, we take a monocular video $\bV \in \mathbb{R}^{T \times H \times W \times 3}$ with $T$ frames as input and extract patch-level features using a pretrained DINOv2~\cite{dinov2} encoder. These semantic features facilitate robust correspondence matching and strong generalization across diverse frames, which is crucial for maintaining consistent motion throughout the sequence. Additionally, we inject temporal embeddings into the patch tokens to make them explicitly aware of frame ordering.

To reconstruct the motion for each frame, a straightforward approach is to represent the entire motion as a single 1D latent sequence and decode frame-wise tokens using positional and temporal queries. While this design is efficient for motions with a fixed length, it cannot naturally handle motion sequences of arbitrary duration. 
To address this limitation, we propose to append the global shape token $\bZ_{\bX_0}$ to each frame token as the frame-wise motion representation, enabling flexible input lengths, as shown in~\figref{fig:pipeline}. Furthermore, we add a reference positional token to explicitly distinguish the reference frame from the others, ensuring that the attention mechanism can properly leverage the reference information during propagation.

With this design, we derive the final latent representation $\bZ_t$ for each frame $t$, which jointly encodes the shared geometric structure and frame-specific motion information.
To distinguish motion features across frames, inspired by VGGT~\cite{VGGT}, we adopt an Alternating-Attention architecture. Let $\mathbf{Z}_t^{(0)} \in \mathbb{R}^{(K+P) \times C}$ denote the initial aggregated shape and visual latent representation for frame $t$. For $L$ alternating attention blocks, the updates are defined as:
\begin{align}
& \textbf{(Global update)} \notag \\
& [\mathbf{Z}_0^{(\ell-\frac{1}{2})}, \dots, \mathbf{Z}_{T-1}^{(\ell-\frac{1}{2})}] 
  {=} \texttt{GlobalAttn}(\mathbf{Z}_0^{(\ell-1)}, \dots, \mathbf{Z}_{T-1}^{(\ell-1)}) \notag \\
& \textbf{(Frame-wise update)} \notag \\
& \mathbf{Z}_t^{(\ell)} 
  = \texttt{FrameAttn}(\mathbf{Z}_t^{(\ell-\frac{1}{2})}), 
  \quad \forall\, t = 0,\dots,T-1
\label{eq:alt-attn}
\end{align}

After $L$ blocks, we obtain the final motion-aware representation for each frame as $\mathbf{Z}_t = \mathbf{Z}_t^{(L)}$. We take the first $K$ tokens as per-frame motion representation.

\subsection{Motion decoding}
\label{sec:motion_decoding}

With the learned shape and motion latent representations, we decode them into explicit per-frame 3D point flows.  
Rather than predicting the full shape independently per-instance~\cite{ShapeGen4D} or attribute offsets per-frame~\cite{gvfd}, we predict per-frame motion flows relative to the reference shape, which preserves surface correspondences and ensures temporal consistency over long sequences.

To achieve this, we adopt a cross-attention decoder: we take a set of points sampled from the reference mesh as queries and predict their positions at each time step.  
Specifically, we resample $M$ points from the reference mesh $\hat{\bP}_0 = \{ (\mathbf{x}_i, \mathbf{n}_i, \mathbf{c}_i) \}_{i=1}^M$.  
These points are embedded using the same \texttt{PointEmb} module employed in shape encoder (\secref{sec:motion_latent_learning}).  

The motion decoder then predicts the per-frame positions independently:
\begin{equation}
\hat{\mathbf{X}}_t = \texttt{MotionDecoder}(\hat{\bX}_0, \mathbf{Z}_t)
\end{equation}
where $\mathbf{Z}_t$ is the motion-aware latent for frame $t$.  

This design enables motion prediction at any spatial location and arbitrary time step, providing a flexible and fully feed-forward 4D reconstruction framework. The decoded point features are subsequently processed through a shared fully connected layer to predict the final 3D coordinates.

\subsection{Training}
We train \method{} with straightforward direct supervision by minimizing the mean squared error (MSE) between the predicted and ground-truth point positions,
\begin{equation}
\mathcal{L} = \frac{1}{M \times T} \sum_{i=1}^{M} \sum_{t=1}^{T} \|\hat{\bX}_t^i - \bX_t^i\|_2^2
\end{equation}

\noindent Here, $\bX_t$ and $\hat{\bX}_t$ denote the ground-truth and predicted point positions, respectively. To capture continuous motion fields, we densely sample points during training. This dense supervision encourages the model to learn fine-grained shape correspondences and ensures coherent motion across the entire mesh.

\section{Experiments}
\subsection{Implementation Details}
\boldparagraph{Dataset.} We curate a high-quality animation dataset by filtering 16,000 objects from a pool of approximately 50,000 models sourced from Objaverse~\cite{Objaverse} and Objaverse-XL~\cite{Objaverse-xl}. Our filtering policy excludes objects with simplistic geometry (e.g., cubes, spheres) and employs Iterative Closest Point (ICP) analysis to discard sequences exhibiting trivial motion. To ensure scale consistency, we normalize each object to fit within a bounding cube defined by the dimensions $[-0.5,0.5]$. For the video data, we render each asset at a resolution of $256\times256$ from fixed viewpoints with uniformly sampled azimuth angles. Both the curated assets and their renderings will be publicly released.

\boldparagraph{Training Strategy.} Our model is trained on 12-frame sequences. For each mesh, we sample $N = 4096 $ points as the shape input, which are encoded into $K = 64$ shape latents. These latents are processed through $L = 16$ transformer blocks to produce motion latents for decoding, and $M = 4096$ for densely sampled ground-truth points. Training is conducted with a total batch size of 256 using 8 H100 GPUs, using a learning rate of $4\times{10}^{-4}$. The model is trained with $60,000$ steps in roughly $1.5$ days.

\subsection{Evaluation}
\begin{table*}[t]
\centering
\renewcommand{\arraystretch}{1.25}
\renewcommand{\tabcolsep}{5pt}
\resizebox{\linewidth}{!}{
\begin{tabular}{l||cc|cccc||cc|cccc||}
  & \multicolumn{6}{c||}{\textbf{Short Sequence}} & \multicolumn{6}{c||}{\textbf{Long Sequence}} \\[2pt]
  \shline
  & \multicolumn{2}{c|}{\textbf{Geometry}} 
  & \multicolumn{4}{c||}{\textbf{Appearance}} 
  & \multicolumn{2}{c|}{\textbf{Geometry}} 
  & \multicolumn{4}{c||}{\textbf{Appearance}} \\[2pt]
  \hline
  \textbf{Method} 
  & {CD $\downarrow$} & {F-Score $\uparrow$} 
  & {LPIPS $\downarrow$} & {CLIP $\uparrow$} & {FVD $\downarrow$} & {DreamSim $\downarrow$}
  & {CD $\downarrow$} & {F-Score $\uparrow$} 
  & {LPIPS $\downarrow$} & {CLIP $\uparrow$} & {FVD $\downarrow$} & {DreamSim $\downarrow$} \\ 
  \hline

  L4GM~\citep{L4gm}  
  & 0.3561 & 0.1269 & \textbf{0.1487} & 0.8182 & \textbf{1120.67} & 0.1941
  & 0.3648 & 0.0997 & \textbf{0.1467} & 0.7988 & \textbf{1070.72} & 0.2175 \\

  GVFD~\citep{gvfd}  
  & 0.1970 & 0.2608 & 0.1664 & 0.7933 & 1414.21 & 0.2147
  & OOM & OOM & OOM & OOM & OOM & OOM \\

  V2M4~\citep{V2M4}  
  & 0.3437 & 0.2318 & 0.1769 & 0.8080 & 1516.47 & 0.1974
  & 0.3719 & 0.1652 & 0.2031 & 0.7872 & 1534.16 & 0.2292 \\

  \textbf{Ours}  
  & \textbf{0.1113} & \textbf{0.3171} & 0.1495 & \textbf{0.8428} & 1175.89 & \textbf{0.1682}
  & \textbf{0.1495} & \textbf{0.2347} & 0.1688 & \textbf{0.8352} & 1264.36 & \textbf{0.1967} \\
 \hline
  \textbf{Ours w/m}  
  & \underline{0.0437} & \underline{0.6774} & \underline{0.0921} & \underline{0.9251} & \underline{497.43} & \underline{0.0614}
  & \underline{0.0929} & \underline{0.4322} & \underline{0.1057} & \underline{0.9224} & \underline{673.03} & \underline{0.0781} \\
\end{tabular}
}
\vspace{-6pt}
\caption{\textbf{Quantitative evaluation on our Motion-80 set.}
Results are reported for both short and long sequences.
“Ours w/m” denotes our method initialized with the ground-truth static mesh from the first frame.
Thanks to the disentangled mesh representation and scene-flow–based motion modeling, our approach capable of transforming artist-created static 3D meshes into fully dynamic 4D sequences.}
\label{tab:mesh}
\end{table*}

\boldparagraph{Evaluation Datasets.} We evaluate our method on two datasets. 
(1) We collect a held-out set of 80 subjects from Objaverse, termed Motion-80, featuring objects with rich textures and diverse motion, which contains 64 short sequences and 16 long sequences exceeding 128 frames rendered from four orthogonal views.
(2) The Consistent4D benchmark~\cite{Consistent4d}, which includes 7 videos of 32 frames each. Since ground-truth meshes are not available for the Consistent4D dataset, we report rendering-based metrics computed at four target novel views following the evaluation protocol in Consistent4D.

\boldparagraph{Baselines.} We conduct comprehensive comparisons with state-of-the-art video-to-4D methods. Our baselines include feedforward approaches that predict 3D Gaussians, \ie, L4GM~\cite{L4gm} and GVFD~\cite{gvfd}.
We also evaluate against the optimization-based method V2M4~\cite{V2M4}, which first generates a 3D mesh for each frame and then performs temporal alignment to obtain a 4D mesh.

\boldparagraph{Metrics.} 
For geometric evaluation, we follow the protocol of Shape2VecSet~\cite{3dshape2vecset} to compute the Chamfer Distance (CD) and F-Score. This involves sampling 50,000 points from each mesh surface for comparison. However, due to the inherent ambiguity of 4D synthesis from a monocular view, the scale and orientation of the generated meshes may not align with the ground truth. To address this, we apply an Iterative Closest Point (ICP) algorithm for alignment prior to evaluation. Specifically, we register the first frame of the generated sequence to the ground-truth mesh to estimate a rigid transformation (rotation, scale, and translation), which is then applied to all subsequent frames in the sequence. For Gaussian-based methods, we extract the centers of all Gaussians as the surface point set and uniformly sample 50,000 points from it to ensure a fair comparison.

For appearance evaluation, we render the textured mesh from target viewpoints. For the Motion-80 evaluation set, we use the front view as input and the remaining 3 views for eval. For the Consistent4D dataset, we follow its original camera settings. We adopt LPIPS~\cite{LPIPS}, CLIP~\cite{clip}, FVD~\cite{FVD}, and DreamSim~\cite{dreamsim} to assess overall quality and temporal consistency, which are widely used in video-to-4D tasks.
We adopt the V2M4~\cite{V2M4} evaluation protocol and further extend it by assessing performance from novel views instead of the input viewpoint.
This provides a more comprehensive evaluation, as it avoids cases where a method appears satisfactory from the input view but exhibits artifacts from other views.

\begin{table}[t]
\centering
\renewcommand{\arraystretch}{1.2}
\renewcommand{\tabcolsep}{10.0pt}
\resizebox{\linewidth}{!}{
\begin{tabular}{l|cccc} 

\textbf{Method} & {LPIPS $\downarrow$} & {CLIP $\uparrow$} & {FVD $\downarrow$} & {DreamSim $\downarrow$} \\ 
\shline

L4GM~\citep{L4gm}  & 0.1468 & 0.8457 & \textbf{1207.79} & 0.1830 \\
GVFD~\citep{gvfd}   & 0.1789 & 0.8278 & 1340.78 & 0.2009 \\
V2M4~\citep{V2M4}   & 0.1611 & 0.8482 & 1471.58 & 0.1832 \\

\bf Ours   & \textbf{0.1455} & \textbf{0.8609} & 1260.06 & \textbf{0.1691} \\

\end{tabular}
}
\vspace{-6pt}
\caption{\textbf{Quantitative evaluation on Consist4D benchmark.} We evaluate rendering performance across 7 test cases, each containing 32 frames, rendered from 4 target novel views.}
\vspace{-10pt}
\label{tab:consist4d}
\end{table}

\begin{figure}[t]
  \centering
    \includegraphics[width=\linewidth]{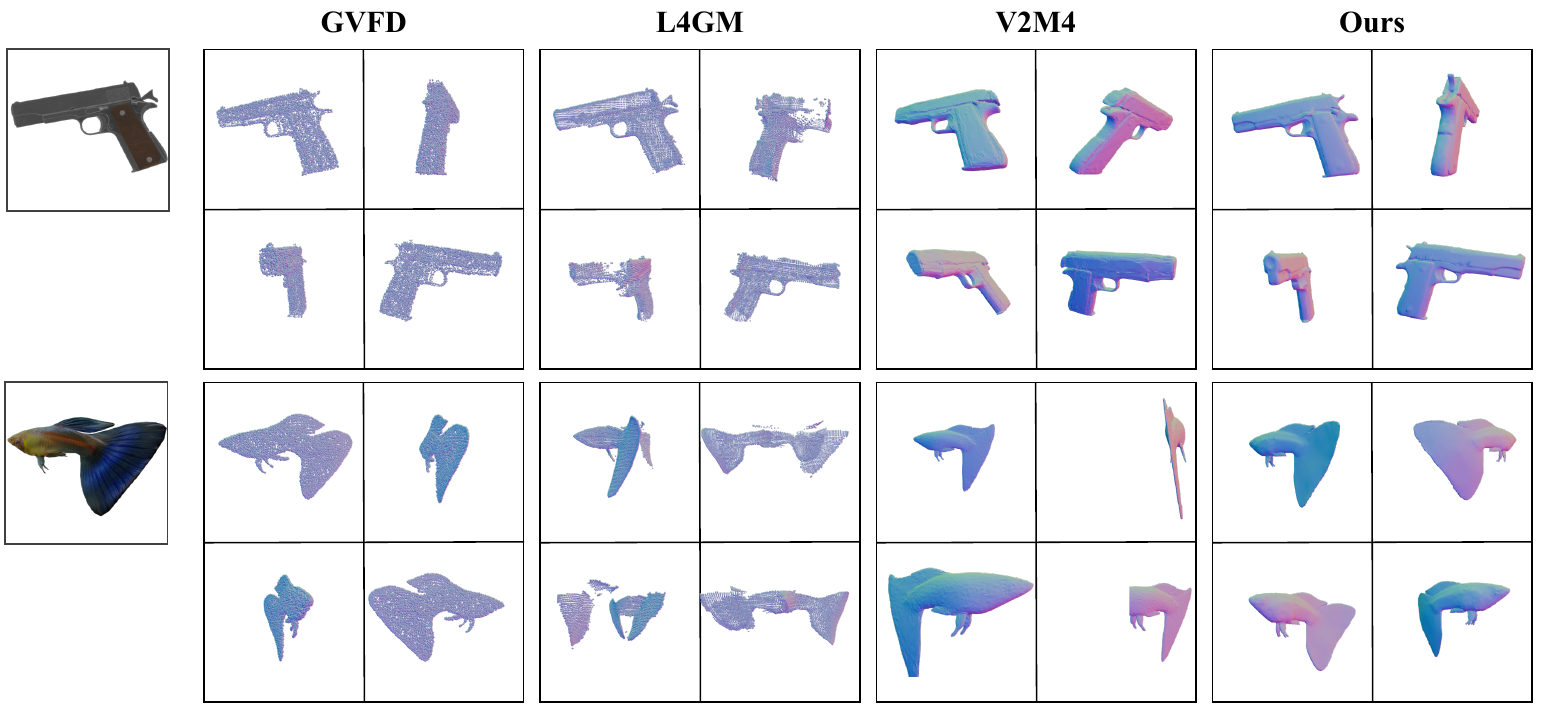}
    \vspace{-15pt}
  \caption{\textbf{Geometric comparison} on the Consistent4D benchmark~\cite{Consistent4d}. Through spatially consistent motion reconstruction, we obtain plausible and high-quality 3D geometry.}
  \label{fig:consist4d}
\end{figure}

\boldparagraph{Quantitative Evaluation}. 
\textit{On geometry.}
As demonstrated in~\tabref{tab:mesh} and Figure~\ref{fig:consist4d}, our method achieves superior 4D geometric accuracy compared to all baselines, as indicated by consistent gain across both CD and F-Score metrics.
Among Gaussian-based methods, L4GM fails to recover accurate geometry. Because the 3DGS representation is tailored for novel-view synthesis, the predicted Gaussians are not constrained to lie exactly on the surface \cite{2dgs}, and the resulting point clouds exhibit floating artifacts, as shown in Figure \ref{fig:consist4d}. GVFD can generate reasonable surface point clouds. However, it struggles to accurately reconstruct the motion of these points over time, leading to degraded overall performance.
The optimization-based method V2M4 refines the generated 3D mesh from each frame and reconstructs a plausible surface, thereby outperforming Gaussian-based approaches. Nonetheless, it still suffers from temporal inconsistency, leading to flickering and physically implausible spatial motion, resulting in low CD scores.
In contrast, we animate the generated explicit 3D mesh with the reconstructed scene flow, producing temporally coherent motion through alignment between the mesh and video observations and thus achieving high geometric accuracy.
This advantage is particularly evident in “Ours w/m”, which drives a ground-truth static mesh from the first frame using our reconstructed motion and significantly outperforms all baselines, highlighting the fidelity of our motion reconstruction.
Notably, \method is the only approach capable of converting artist-created static 3D meshes into dynamic 4D sequences, owing to its disentangled mesh representation and scene-flow–based motion modeling.

\begin{figure*}[t]
  \centering
  \includegraphics[width=0.98\linewidth]{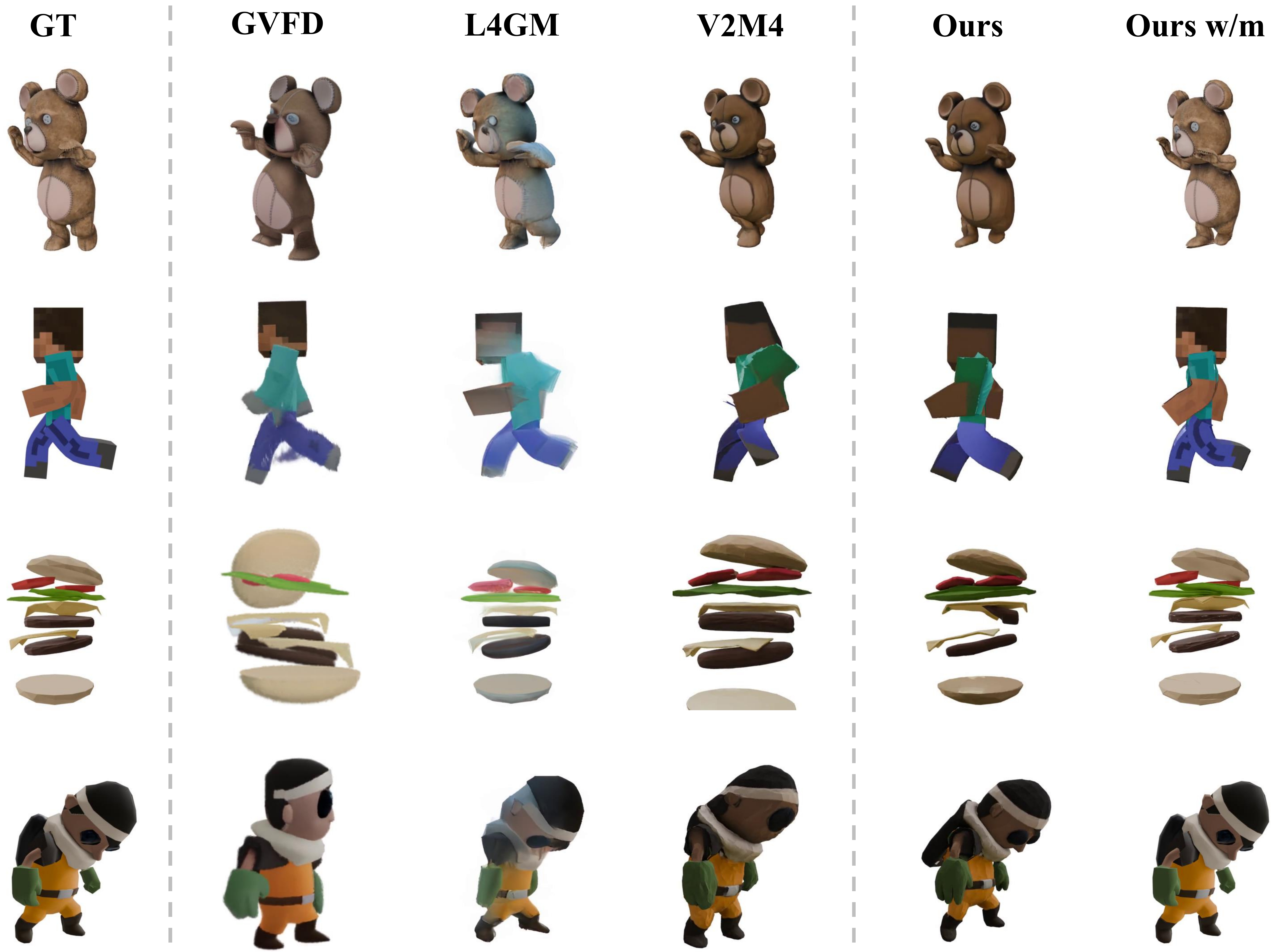}
  \caption{\textbf{Qualitative Comparisons.} We compare our method with strong baselines including GVFD~\cite{gvfd}, L4GM~\cite{L4gm}, and V2M4~\cite{V2M4} on our proposed Motion-80 benchmark. For fair evaluation, we render the generated 4D assets from all methods into an orthogonal novel view. Our approach produces more temporally coherent and structurally consistent motion. We invite reviewers to consult the supplemental material for animation visualization.}
  \label{fig:objaverse}
\end{figure*}
\textit{On appearance.} 
As shown in \tabref{tab:mesh} and \tabref{tab:consist4d}, our approach achieves better 3D content fidelity and consistency, quantitatively outperforms baselines in CLIP and DreamSim metrics. 
Note that L4GM is trained and evaluated on the orthogonal view, thus its rendering is biased to the evaluation protocol and may have more advantages than our method on the specific rendering perspective.
However, when viewed from non-orthogonal novel viewpoints, L4GM exhibits noticeable ghosting artifacts, whereas our method continues to produce plausible and stable results. 
We invite the reviewer to our supplemental material for more results.

\begin{figure*}[h]
  \centering
  \includegraphics[width=1.0\linewidth]{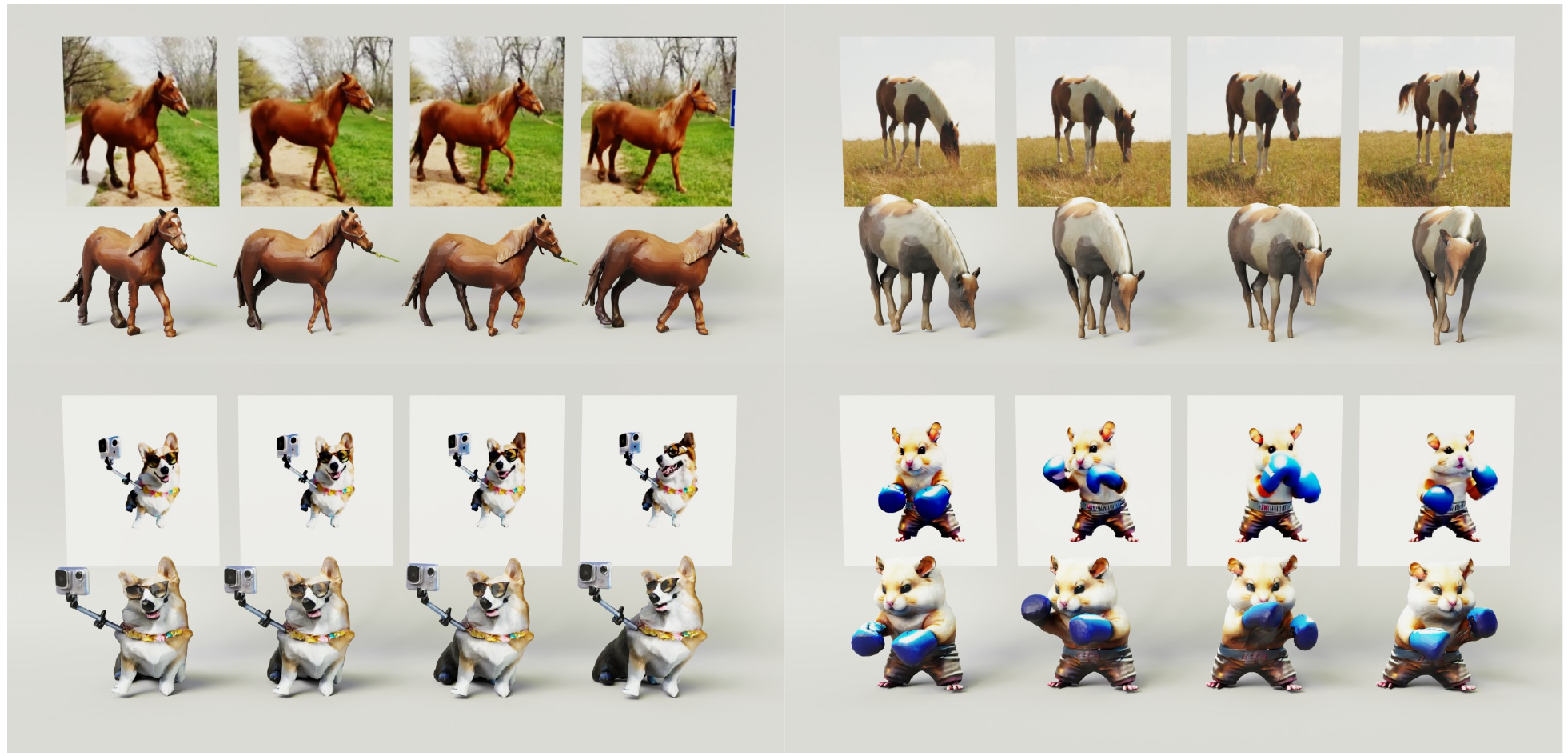}
  \vspace{-3mm}
  \caption{\textbf{In-the-Wild Video-to-4D Synthesis.}
  Our method generalizes to diverse in-the-wild inputs, including real-world videos (top row) and generated animations (bottom row). By formulating motion reconstruction as surface-to-pixel alignment, we achieve robust local correspondence reasoning across varied shapes and motion patterns.
  }
  \label{fig:wild4d}
\end{figure*}

\boldparagraph{Qualitative Evaluation.} 
 \figref{fig:objaverse} further illustrates the benefits of our reformulation.
L4GM suffers from error accumulation during multi-view generation, resulting in ghosting artifacts when viewed from angles different from the input views.
GVFD generates jittery, temporally inconsistent motions due to limitations in its VAE-based motion modeling that relies on large datasets to learn motion latent distribution, leading to weak generalization and discontinuous appearance in the Gaussian during movement.
V2M4, relying on per-frame optimization, generates plausible results from the input view, but suffers from spatial discontinuities when observed from other viewpoints, failing to capture true motion. 
In contrast, our method combines a strong pretrained 3D generator with feed-forward, generalizable motion reconstruction. This design preserves both spatial and temporal coherence, produces smooth, physically plausible motion, and achieves superior visual fidelity.

\subsection{More Results}

\boldparagraph{Wild4D.} We show a few in-the-wild testing samples~\figref{fig:wild4d}. To handle out-of-domain inputs, we first apply BiRefNet~\cite{biref} to automatically remove background regions on a per-frame basis. We then generate an initial 3D shape using the first video frame with Hunyuan2.0~\cite{Hunyuan3d_2}.
Thanks to the strong visual features from the DINO encoder and the robust geometry encoder, our method generalizes well to in-the-wild video inputs, including both real-world footage and generated animation sequences.

\begin{figure}[h]
  \centering
  \includegraphics[width=\linewidth]{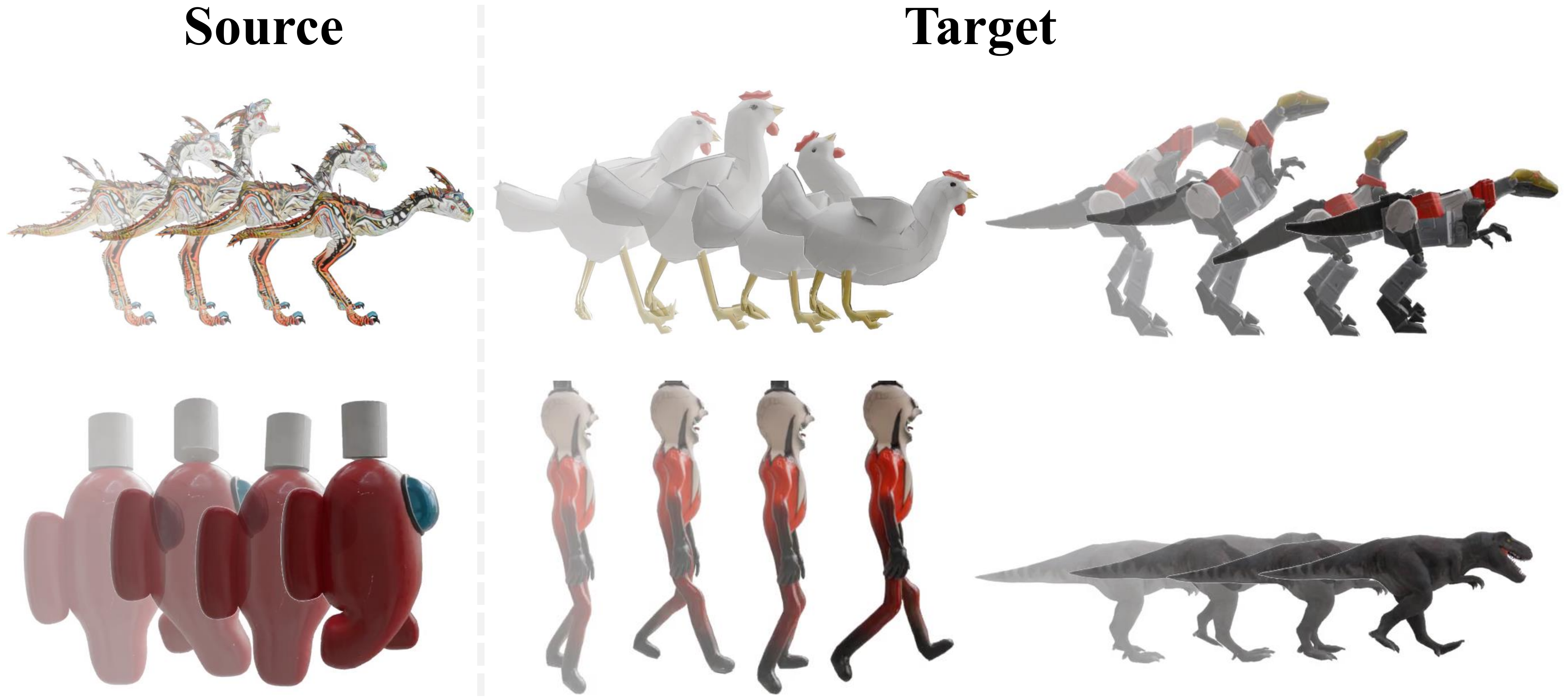}
  \caption{\textbf{Motion Transfer Example.} By disentangling 4D synthesis into 3D mesh generation and motion reconstruction, our framework can animate static articulated objects with motion retargeted from videos of different sources.}
  \label{fig:motion_trans}
  \vspace{-3mm}
\end{figure}

\boldparagraph{Motion Transfer.} Although our model is trained using paired videos and corresponding 3D subjects, we observe that it inherently generalizes to transferring the motion from an input video to a 3D object with different shape and appearance. As shown in~\figref{fig:motion_trans}, we feed the dragon video into the DINO encoder and use the chicken and robot-dragon meshes as inputs to the geometry encoder. Remarkably, our method successfully transfers the neck, body and leg motions from the video to these new target objects. For shapes with geometric differences, our method is still able to successfully transfer the leg movement from the source to the target model.

\section{Conclusion}
We presented a 4D synthesis pipeline that decomposes the inherently ill-posed 4D generation problem into two tractable components: 3D shape generation and motion reconstruction. This decomposition offers several key advantages. \textit{Efficiency:} Off-the-shelf 3D generative models can be directly reused for high-quality shape synthesis, while the motion branch remains lightweight, substantially reducing the scale requirements for 4D training data. \textit{Generalizability:} By formulating motion reconstruction as an alignment task between surface points and video pixels, our method performs robust local correspondence reasoning, enabling strong generalization to both synthesized and real-world shapes as well as diverse motion patterns. \textit{Flexibility:} Our framework also supports lifting static articulated objects into dynamic 4D driven by video conditions, including motion retargeting from entirely different sources.

\begin{figure}[t]
  \centering
  \includegraphics[width=\linewidth]{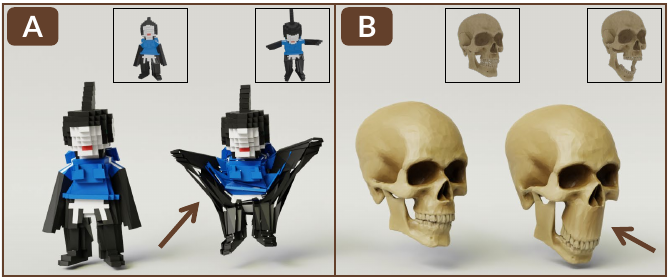}
  \caption{\textbf{Failure cases.} (A) Vertex sticking in challenging cases. (B) Initial mesh topology not able to adapt to later motion.}
  \label{fig:fail}
\end{figure}
\vspace{1mm}
\boldparagraph{Limitations.} Our method exhibits several limitations. First, the geometry encoder operates on a dense point cloud without explicitly modeling the mesh topology. As a result, when different parts of the object are not clearly separated in the reference mesh, the model may produce vertex sticking artifacts, as illustrated in \figref{fig:fail}(A). Second, when reconstructing motion from monocular video, our pipeline relies on the mesh generated from the first frame as reference geometry. This makes it difficult to accommodate topology changes that occur in later frames, leading to failure cases, as shown in \figref{fig:fail}(B).

\clearpage
\section*{Acknowledgments}
We would like to thank the members of \textit{Inception3D} Lab for their helpful discussions, and \textit{Isabella Liu} for sharing the Blender scripts used in rendering the teaser. This work is done with the sponsorship of TeleAI.\par

{
    \small
    \bibliographystyle{ieeenat_fullname}
    \bibliography{bibliography,bibliography_short,main}

\begin{thebibliography}{113}
\providecommand{\natexlab}[1]{#1}
\providecommand{\url}[1]{\texttt{#1}}
\expandafter\ifx\csname urlstyle\endcsname\relax
  \providecommand{\doi}[1]{doi: #1}\else
  \providecommand{\doi}{doi: \begingroup \urlstyle{rm}\Url}\fi

\bibitem[Agarwal et~al.(2011)Agarwal, Furukawa, Snavely, Simon, Curless, Seitz, and Szeliski]{agarwal2011building}
Sameer Agarwal, Yasutaka Furukawa, Noah Snavely, Ian Simon, Brian Curless, Steven~M Seitz, and Richard Szeliski.
\newblock Building rome in a day.
\newblock \emph{ACM Communications}, 2011.

\bibitem[Bahmani et~al.(2024)Bahmani, Skorokhodov, Rong, Wetzstein, Guibas, Wonka, Tulyakov, Park, Tagliasacchi, and Lindell]{4d-fy}
Sherwin Bahmani, Ivan Skorokhodov, Victor Rong, Gordon Wetzstein, Leonidas Guibas, Peter Wonka, Sergey Tulyakov, Jeong~Joon Park, Andrea Tagliasacchi, and David~B Lindell.
\newblock 4d-fy: Text-to-4d generation using hybrid score distillation sampling.
\newblock In \emph{CVPR}, 2024.

\bibitem[Baran and Popovi{\'c}(2007)]{Pinocchio}
Ilya Baran and Jovan Popovi{\'c}.
\newblock Automatic rigging and animation of 3d characters.
\newblock \emph{ACM Trans. on Graphics}, 2007.

\bibitem[Bekor et~al.(2025)Bekor, Harari, Perel, and Litany]{gaussiansee}
Yarin Bekor, Gal~Michael Harari, Or Perel, and Or Litany.
\newblock Gaussian see, gaussian do: Semantic 3d motion transfer from multiview video.
\newblock In \emph{SIGGRAPH Asia}, pages 1--10, 2025.

\bibitem[Chan et~al.(2022)Chan, Lin, Chan, Nagano, Pan, De~Mello, Gallo, Guibas, Tremblay, Khamis, et~al.]{eg3d}
Eric~R Chan, Connor~Z Lin, Matthew~A Chan, Koki Nagano, Boxiao Pan, Shalini De~Mello, Orazio Gallo, Leonidas~J Guibas, Jonathan Tremblay, Sameh Khamis, et~al.
\newblock Efficient geometry-aware 3d generative adversarial networks.
\newblock In \emph{CVPR}, 2022.

\bibitem[Chen et~al.(2022)Chen, Liu, Xie, Chen, Su, and Yu]{sofgan}
Anpei Chen, Ruiyang Liu, Ling Xie, Zhang Chen, Hao Su, and Jingyi Yu.
\newblock Sofgan: A portrait image generator with dynamic styling.
\newblock \emph{ACM Trans. on Graphics}, 2022.

\bibitem[Chen et~al.(2025{\natexlab{a}})Chen, Zhang, Tang, and Wonka]{V2M4}
Jianqi Chen, Biao Zhang, Xiangjun Tang, and Peter Wonka.
\newblock V2m4: 4d mesh animation reconstruction from a single monocular video.
\newblock In \emph{ICCV}, 2025{\natexlab{a}}.

\bibitem[Chen et~al.(2025{\natexlab{b}})Chen, Zhang, Yin, Dou, Chen, Wang, Komura, and Zhang]{Motion2Motion}
Ling-Hao Chen, Yuhong Zhang, Zixin Yin, Zhiyang Dou, Xin Chen, Jingbo Wang, Taku Komura, and Lei Zhang.
\newblock Motion2motion: Cross-topology motion transfer with sparse correspondence.
\newblock In \emph{SIGGRAPH Asia}, 2025{\natexlab{b}}.

\bibitem[Chen et~al.(2025{\natexlab{c}})Chen, Chen, Xiu, Geiger, and Chen]{TTT3R}
Xingyu Chen, Yue Chen, Yuliang Xiu, Andreas Geiger, and Anpei Chen.
\newblock Ttt3r: 3d reconstruction as test-time training.
\newblock \emph{arXiv preprint arXiv:2509.26645}, 2025{\natexlab{c}}.

\bibitem[Chen et~al.(2025{\natexlab{d}})Chen, Chen, Xiu, Geiger, and Chen]{easi3r}
Xingyu Chen, Yue Chen, Yuliang Xiu, Andreas Geiger, and Anpei Chen.
\newblock Easi3r: Estimating disentangled motion from dust3r without training.
\newblock In \emph{ICCV}, 2025{\natexlab{d}}.

\bibitem[Davison et~al.(2007)Davison, Reid, Molton, and Stasse]{davison2007monoslam}
Andrew~J Davison, Ian~D Reid, Nicholas~D Molton, and Olivier Stasse.
\newblock Monoslam: Real-time single camera slam.
\newblock \emph{PAMI}, 2007.

\bibitem[Deitke et~al.(2023{\natexlab{a}})Deitke, Liu, Wallingford, Ngo, Michel, Kusupati, Fan, Laforte, Voleti, Gadre, et~al.]{Objaverse-xl}
Matt Deitke, Ruoshi Liu, Matthew Wallingford, Huong Ngo, Oscar Michel, Aditya Kusupati, Alan Fan, Christian Laforte, Vikram Voleti, Samir~Yitzhak Gadre, et~al.
\newblock Objaverse-xl: A universe of 10m+ 3d objects.
\newblock \emph{NeurIPS}, 2023{\natexlab{a}}.

\bibitem[Deitke et~al.(2023{\natexlab{b}})Deitke, Schwenk, Salvador, Weihs, Michel, VanderBilt, Schmidt, Ehsani, Kembhavi, and Farhadi]{Objaverse}
Matt Deitke, Dustin Schwenk, Jordi Salvador, Luca Weihs, Oscar Michel, Eli VanderBilt, Ludwig Schmidt, Kiana Ehsani, Aniruddha Kembhavi, and Ali Farhadi.
\newblock Objaverse: A universe of annotated 3d objects.
\newblock In \emph{CVPR}, 2023{\natexlab{b}}.

\bibitem[Deng et~al.(2025)Deng, Zhang, Geng, Wu, and Wu]{Anymate}
Yufan Deng, Yuhao Zhang, Chen Geng, Shangzhe Wu, and Jiajun Wu.
\newblock Anymate: A dataset and baselines for learning 3d object rigging.
\newblock In \emph{SIGGRAPH}, 2025.

\bibitem[Feng et~al.(2025)Feng, Zhang, Wang, Ye, Yu, Black, Darrell, and Kanazawa]{St4rtrack}
Haiwen Feng, Junyi Zhang, Qianqian Wang, Yufei Ye, Pengcheng Yu, Michael~J Black, Trevor Darrell, and Angjoo Kanazawa.
\newblock St4rtrack: Simultaneous 4d reconstruction and tracking in the world.
\newblock In \emph{ICCV}, 2025.

\bibitem[Fu et~al.(2023)Fu, Tamir, Sundaram, Chai, Zhang, Dekel, and Isola]{dreamsim}
Stephanie Fu, Netanel~Y Tamir, Shobhita Sundaram, Lucy Chai, Richard Zhang, Tali Dekel, and Phillip Isola.
\newblock Dreamsim: learning new dimensions of human visual similarity using synthetic data.
\newblock \emph{NeurIPS}, 2023.

\bibitem[Gao et~al.(2024)Gao, Liu, Chen, Geiger, and Schölkopf]{gao2024graphdreamer}
Gege Gao, Weiyang Liu, Anpei Chen, Andreas Geiger, and Bernhard Schölkopf.
\newblock Graphdreamer: Compositional 3d scene synthesis from scene graphs.
\newblock In \emph{CVPR}, 2024.

\bibitem[Gao et~al.(2022)Gao, Shen, Wang, Chen, Yin, Li, Litany, Gojcic, and Fidler]{get3d}
Jun Gao, Tianchang Shen, Zian Wang, Wenzheng Chen, Kangxue Yin, Daiqing Li, Or Litany, Zan Gojcic, and Sanja Fidler.
\newblock Get3d: A generative model of high quality 3d textured shapes learned from images.
\newblock \emph{NeurIPS}, 2022.

\bibitem[Guo et~al.(2025{\natexlab{a}})Guo, Xiang, Ma, Zhou, Li, and Zhang]{makeani}
Zhiyang Guo, Jinxu Xiang, Kai Ma, Wengang Zhou, Houqiang Li, and Ran Zhang.
\newblock Make-it-animatable: An efficient framework for authoring animation-ready 3d characters.
\newblock In \emph{CVPR}, 2025{\natexlab{a}}.

\bibitem[Guo et~al.(2025{\natexlab{b}})Guo, Zhang, Xiang, Zhao, Zhou, and Li]{makepose}
Zhiyang Guo, Ori Zhang, Jax Xiang, Alan Zhao, Wengang Zhou, and Houqiang Li.
\newblock Make-it-poseable: Feed-forward latent posing model for 3d humanoid character animation.
\newblock \emph{arXiv preprint arXiv:2512.16767}, 2025{\natexlab{b}}.

\bibitem[He et~al.(2025)He, Geng, Wu, and Wu]{he2025category}
Guangzhao He, Chen Geng, Shangzhe Wu, and Jiajun Wu.
\newblock Category-agnostic neural object rigging.
\newblock In \emph{CVPR}, 2025.

\bibitem[He et~al.(2017)He, Gkioxari, Dollar, and Girshick]{MaskRCNN}
Kaiming He, Georgia Gkioxari, Piotr Dollar, and Ross Girshick.
\newblock Mask {R-CNN}.
\newblock In \emph{ICCV}, 2017.

\bibitem[Hu et~al.(2025)Hu, Gao, Li, Zhao, Cun, Zhang, Quan, and Shan]{hu2024depthcrafter}
Wenbo Hu, Xiangjun Gao, Xiaoyu Li, Sijie Zhao, Xiaodong Cun, Yong Zhang, Long Quan, and Ying Shan.
\newblock Depthcrafter: Generating consistent long depth sequences for open-world videos.
\newblock In \emph{CVPR}, 2025.

\bibitem[Huang et~al.(2024)Huang, Yu, Chen, Geiger, and Gao]{2dgs}
Binbin Huang, Zehao Yu, Anpei Chen, Andreas Geiger, and Shenghua Gao.
\newblock 2d gaussian splatting for geometrically accurate radiance fields.
\newblock In \emph{SIGGRAPH}, 2024.

\bibitem[Huang et~al.(2025{\natexlab{a}})Huang, Feng, Sun, Guo, Cao, and Sheng]{animax_video}
Zehuan Huang, Haoran Feng, Yangtian Sun, Yuanchen Guo, Yanpei Cao, and Lu Sheng.
\newblock Animax: Animating the inanimate in 3d with joint video-pose diffusion models.
\newblock \emph{arXiv preprint arXiv:2506.19851}, 2025{\natexlab{a}}.

\bibitem[Huang et~al.(2025{\natexlab{b}})Huang, Guo, Wang, Yi, Ma, Cao, and Sheng]{Mv-adapter}
Zehuan Huang, Yuan-Chen Guo, Haoran Wang, Ran Yi, Lizhuang Ma, Yan-Pei Cao, and Lu Sheng.
\newblock Mv-adapter: Multi-view consistent image generation made easy.
\newblock In \emph{ICCV}, 2025{\natexlab{b}}.

\bibitem[Jain et~al.(2022)Jain, Mildenhall, Barron, Abbeel, and Poole]{DreamFields}
Ajay Jain, Ben Mildenhall, Jonathan~T Barron, Pieter Abbeel, and Ben Poole.
\newblock Zero-shot text-guided object generation with dream fields.
\newblock In \emph{CVPR}, 2022.

\bibitem[Jiang et~al.(2024{\natexlab{a}})Jiang, Yu, Cao, Wang, Hu, and Gao]{jiang2024animate3d}
Yanqin Jiang, Chaohui Yu, Chenjie Cao, Fan Wang, Weiming Hu, and Jin Gao.
\newblock Animate3d: Animating any 3d model with multi-view video diffusion.
\newblock \emph{NeurIPS}, 2024{\natexlab{a}}.

\bibitem[Jiang et~al.(2024{\natexlab{b}})Jiang, Zhang, Gao, Hu, and Yao]{Consistent4d}
Yanqin Jiang, Li Zhang, Jin Gao, Weimin Hu, and Yao Yao.
\newblock Consistent4d: Consistent 360$^\circ$ dynamic object generation from monocular video.
\newblock In \emph{ICLR}, 2024{\natexlab{b}}.

\bibitem[Kerbl et~al.(2023)Kerbl, Kopanas, Leimk{\"u}hler, and Drettakis]{3dgs}
Bernhard Kerbl, Georgios Kopanas, Thomas Leimk{\"u}hler, and George Drettakis.
\newblock 3d gaussian splatting for real-time radiance field rendering.
\newblock \emph{ACM Trans. on Graphics}, 2023.

\bibitem[Kopf et~al.(2021)Kopf, Rong, and Huang]{Robust-CVD}
Johannes Kopf, Xuejian Rong, and Jia-Bin Huang.
\newblock Robust consistent video depth estimation.
\newblock In \emph{CVPR}, 2021.

\bibitem[Lai et~al.(2025)Lai, Zhao, Liu, Zhao, Lin, Shi, Yang, Yang, Yang, Feng, et~al.]{Hunyuan3d_25}
Zeqiang Lai, Yunfei Zhao, Haolin Liu, Zibo Zhao, Qingxiang Lin, Huiwen Shi, Xianghui Yang, Mingxin Yang, Shuhui Yang, Yifei Feng, et~al.
\newblock Hunyuan3d 2.5: Towards high-fidelity 3d assets generation with ultimate details.
\newblock \emph{arXiv preprint arXiv:2506.16504}, 2025.

\bibitem[Lei et~al.(2025)Lei, Weng, Harley, Guibas, and Daniilidis]{Mosca}
Jiahui Lei, Yijia Weng, Adam~W Harley, Leonidas Guibas, and Kostas Daniilidis.
\newblock Mosca: Dynamic gaussian fusion from casual videos via 4d motion scaffolds.
\newblock In \emph{CVPR}, 2025.

\bibitem[Li et~al.(2024{\natexlab{a}})Li, Tan, Zhang, Xu, Luan, Xu, Hong, Sunkavalli, Shakhnarovich, and Bi]{Instant3d}
Jiahao Li, Hao Tan, Kai Zhang, Zexiang Xu, Fujun Luan, Yinghao Xu, Yicong Hong, Kalyan Sunkavalli, Greg Shakhnarovich, and Sai Bi.
\newblock Instant3d: Fast text-to-3d with sparse-view generation and large reconstruction model.
\newblock In \emph{ICLR}, 2024{\natexlab{a}}.

\bibitem[Li et~al.(2025{\natexlab{a}})Li, Zhang, Sun, Qi, Li, Cheng, Cai, Wu, Liu, Wang, et~al.]{Step1x-3d}
Weiyu Li, Xuanyang Zhang, Zheng Sun, Di Qi, Hao Li, Wei Cheng, Weiwei Cai, Shihao Wu, Jiarui Liu, Zihao Wang, et~al.
\newblock Step1x-3d: Towards high-fidelity and controllable generation of textured 3d assets.
\newblock \emph{arXiv preprint arXiv:2505.07747}, 2025{\natexlab{a}}.

\bibitem[Li et~al.(2025{\natexlab{b}})Li, Ma, Lin, Chen, Jiang, Liu, and Xiang]{akd_articulatedk_diffusion}
Xuan Li, Qianli Ma, Tsung-Yi Lin, Yongxin Chen, Chenfanfu Jiang, Ming-Yu Liu, and Donglai Xiang.
\newblock Articulated kinematics distillation from video diffusion models.
\newblock In \emph{CVPR}, 2025{\natexlab{b}}.

\bibitem[Li et~al.(2021)Li, Niklaus, Snavely, and Wang]{Neuralsceneflow}
Zhengqi Li, Simon Niklaus, Noah Snavely, and Oliver Wang.
\newblock Neural scene flow fields for space-time view synthesis of dynamic scenes.
\newblock In \emph{CVPR}, 2021.

\bibitem[Li et~al.(2024{\natexlab{b}})Li, Chen, and Liu]{Dreammesh4d}
Zhiqi Li, Yiming Chen, and Peidong Liu.
\newblock Dreammesh4d: Video-to-4d generation with sparse-controlled gaussian-mesh hybrid representation.
\newblock \emph{NeurIPS}, 2024{\natexlab{b}}.

\bibitem[Li et~al.(2024{\natexlab{c}})Li, Litvak, Li, Zhang, Jakab, Rupprecht, Wu, Vedaldi, and Wu]{3dfauna}
Zizhang Li, Dor Litvak, Ruining Li, Yunzhi Zhang, Tomas Jakab, Christian Rupprecht, Shangzhe Wu, Andrea Vedaldi, and Jiajun Wu.
\newblock Learning the 3d fauna of the web.
\newblock In \emph{CVPR}, 2024{\natexlab{c}}.

\bibitem[Li et~al.(2025{\natexlab{c}})Li, Tucker, Cole, Wang, Jin, Ye, Kanazawa, Holynski, and Snavely]{MegaSaM}
Zhengqi Li, Richard Tucker, Forrester Cole, Qianqian Wang, Linyi Jin, Vickie Ye, Angjoo Kanazawa, Aleksander Holynski, and Noah Snavely.
\newblock {MegaSaM:} accurate, fast, and robust structure and motion from casual dynamic videos.
\newblock In \emph{CVPR}, 2025{\natexlab{c}}.

\bibitem[Li et~al.(2025{\natexlab{d}})Li, Wang, Zheng, Luo, and Wen]{Sparc3D}
Zhihao Li, Yufei Wang, Heliang Zheng, Yihao Luo, and Bihan Wen.
\newblock Sparc3d: Sparse representation and construction for high-resolution 3d shapes modeling.
\newblock \emph{arXiv preprint arXiv:2505.14521}, 2025{\natexlab{d}}.

\bibitem[Liang et~al.(2024)Liang, Yin, Xu, Liang, Wang, Plataniotis, Zhao, and Wei]{Diffusion4d}
Hanwen Liang, Yuyang Yin, Dejia Xu, Hanxue Liang, Zhangyang Wang, Konstantinos~N Plataniotis, Yao Zhao, and Yunchao Wei.
\newblock Diffusion4d: Fast spatial-temporal consistent 4d generation via video diffusion models.
\newblock \emph{arXiv preprint arXiv:2405.16645}, 2024.

\bibitem[Liao et~al.(2025)Liao, Liu, Xu, Ge, Yang, and Huang]{pad3r}
Ting-Hsuan Liao, Haowen Liu, Yiran Xu, Songwei Ge, Gengshan Yang, and Jia-Bin Huang.
\newblock Pad3r: Pose-aware dynamic 3d reconstruction from casual videos.
\newblock In \emph{SIGGRAPH Asia}, pages 1--11, 2025.

\bibitem[Lin et~al.(2023)Lin, Gao, Tang, Takikawa, Zeng, Huang, Kreis, Fidler, Liu, and Lin]{Magic3d}
Chen-Hsuan Lin, Jun Gao, Luming Tang, Towaki Takikawa, Xiaohui Zeng, Xun Huang, Karsten Kreis, Sanja Fidler, Ming-Yu Liu, and Tsung-Yi Lin.
\newblock Magic3d: High-resolution text-to-3d content creation.
\newblock In \emph{CVPR}, 2023.

\bibitem[Ling et~al.(2024)Ling, Kim, Torralba, Fidler, and Kreis]{ling2024align}
Huan Ling, Seung~Wook Kim, Antonio Torralba, Sanja Fidler, and Karsten Kreis.
\newblock Align your gaussians: Text-to-4d with dynamic 3d gaussians and composed diffusion models.
\newblock In \emph{CVPR}, 2024.

\bibitem[Liu et~al.(2024{\natexlab{a}})Liu, Su, and Wang]{DG-Mesh}
Isabella Liu, Hao Su, and Xiaolong Wang.
\newblock Dynamic gaussians mesh: Consistent mesh reconstruction from dynamic scenes.
\newblock \emph{arXiv preprint arXiv:2404.12379}, 2024{\natexlab{a}}.

\bibitem[Liu et~al.(2025{\natexlab{a}})Liu, Xu, Yifan, Tan, Xu, Wang, Su, and Shi]{riganything}
Isabella Liu, Zhan Xu, Wang Yifan, Hao Tan, Zexiang Xu, Xiaolong Wang, Hao Su, and Zifan Shi.
\newblock Riganything: Template-free autoregressive rigging for diverse 3d assets.
\newblock \emph{ACM Trans. on Graphics}, 2025{\natexlab{a}}.

\bibitem[Liu et~al.(2023)Liu, Wu, Van~Hoorick, Tokmakov, Zakharov, and Vondrick]{Zero-1-to-3}
Ruoshi Liu, Rundi Wu, Basile Van~Hoorick, Pavel Tokmakov, Sergey Zakharov, and Carl Vondrick.
\newblock Zero-1-to-3: Zero-shot one image to 3d object.
\newblock In \emph{ICCV}, 2023.

\bibitem[Liu et~al.(2025{\natexlab{b}})Liu, Xiao, Chen, Feng, Tai, Tang, and Kang]{trace}
Xinhang Liu, Yuxi Xiao, Donny~Y Chen, Jiashi Feng, Yu-Wing Tai, Chi-Keung Tang, and Bingyi Kang.
\newblock Trace anything: Representing any video in 4d via trajectory fields.
\newblock \emph{arXiv preprint arXiv:2510.13802}, 2025{\natexlab{b}}.

\bibitem[Liu et~al.(2024{\natexlab{b}})Liu, Lin, Zeng, Long, Liu, Komura, and Wang]{Syncdreamer}
Yuan Liu, Cheng Lin, Zijiao Zeng, Xiaoxiao Long, Lingjie Liu, Taku Komura, and Wenping Wang.
\newblock Syncdreamer: Generating multiview-consistent images from a single-view image.
\newblock In \emph{ICLR}, 2024{\natexlab{b}}.

\bibitem[Luo et~al.(2020)Luo, Huang, Szeliski, Matzen, and Kopf]{cvd}
Xuan Luo, Jia-Bin Huang, Richard Szeliski, Kevin Matzen, and Johannes Kopf.
\newblock Consistent video depth estimation.
\newblock \emph{ACM Trans. on Graphics}, 2020.

\bibitem[Ma et~al.(2025)Ma, Chen, Yu, Bi, Zhang, Ziwen, Xu, Yang, Xu, Sunkavalli, et~al.]{4dlrm}
Ziqiao Ma, Xuweiyi Chen, Shoubin Yu, Sai Bi, Kai Zhang, Chen Ziwen, Sihan Xu, Jianing Yang, Zexiang Xu, Kalyan Sunkavalli, et~al.
\newblock 4d-lrm: Large space-time reconstruction model from and to any view at any time.
\newblock \emph{arXiv preprint arXiv:2506.18890}, 2025.

\bibitem[Mildenhall et~al.(2021)Mildenhall, Srinivasan, Tancik, Barron, Ramamoorthi, and Ng]{nerf}
Ben Mildenhall, Pratul~P Srinivasan, Matthew Tancik, Jonathan~T Barron, Ravi Ramamoorthi, and Ren Ng.
\newblock Nerf: Representing scenes as neural radiance fields for view synthesis.
\newblock \emph{Communications of the ACM}, 2021.

\bibitem[Mur-Artal et~al.(2015)Mur-Artal, Montiel, and Tardos]{mur2015orb}
Raul Mur-Artal, Jose Maria~Martinez Montiel, and Juan~D Tardos.
\newblock Orb-slam: a versatile and accurate monocular slam system.
\newblock \emph{IEEE transactions on robotics}, 2015.

\bibitem[Niemeyer and Geiger(2021)]{giraffe}
Michael Niemeyer and Andreas Geiger.
\newblock Giraffe: Representing scenes as compositional generative neural feature fields.
\newblock In \emph{CVPR}, 2021.

\bibitem[Oquab et~al.()Oquab, Darcet, Moutakanni, Vo, Szafraniec, Khalidov, Fernandez, Haziza, Massa, El-Nouby, et~al.]{dinov2}
Maxime Oquab, Timoth{\'e}e Darcet, Th{\'e}o Moutakanni, Huy Vo, Marc Szafraniec, Vasil Khalidov, Pierre Fernandez, Daniel Haziza, Francisco Massa, Alaaeldin El-Nouby, et~al.
\newblock Dinov2: Learning robust visual features without supervision.
\newblock \emph{Transactions on Machine Learning Research}.

\bibitem[Poole et~al.(2023)Poole, Jain, Barron, and Mildenhall]{Dreamfusion}
Ben Poole, Ajay Jain, Jonathan~T Barron, and Ben Mildenhall.
\newblock Dreamfusion: Text-to-3d using 2d diffusion.
\newblock In \emph{ICLR}, 2023.

\bibitem[Qiu et~al.(2024)Qiu, Chen, Gu, Zuo, Xu, Wu, Yuan, Dong, Bo, and Han]{Richdreamer}
Lingteng Qiu, Guanying Chen, Xiaodong Gu, Qi Zuo, Mutian Xu, Yushuang Wu, Weihao Yuan, Zilong Dong, Liefeng Bo, and Xiaoguang Han.
\newblock Richdreamer: A generalizable normal-depth diffusion model for detail richness in text-to-3d.
\newblock In \emph{CVPR}, 2024.

\bibitem[Radford et~al.(2021)Radford, Kim, Hallacy, Ramesh, Goh, Agarwal, Sastry, Askell, Mishkin, Clark, et~al.]{clip}
Alec Radford, Jong~Wook Kim, Chris Hallacy, Aditya Ramesh, Gabriel Goh, Sandhini Agarwal, Girish Sastry, Amanda Askell, Pamela Mishkin, Jack Clark, et~al.
\newblock Learning transferable visual models from natural language supervision.
\newblock In \emph{ICCV}, 2021.

\bibitem[Ren et~al.(2024)Ren, Xie, Mirzaei, Kreis, Liu, Torralba, Fidler, Kim, Ling, et~al.]{L4gm}
Jiawei Ren, Cheng Xie, Ashkan Mirzaei, Karsten Kreis, Ziwei Liu, Antonio Torralba, Sanja Fidler, Seung~Wook Kim, Huan Ling, et~al.
\newblock L4gm: Large 4d gaussian reconstruction model.
\newblock \emph{NeurIPS}, 2024.

\bibitem[Schwarz et~al.(2020)Schwarz, Liao, Niemeyer, and Geiger]{graf}
Katja Schwarz, Yiyi Liao, Michael Niemeyer, and Andreas Geiger.
\newblock Graf: Generative radiance fields for 3d-aware image synthesis.
\newblock \emph{NeurIPS}, 2020.

\bibitem[Schönberger and Frahm(2016)]{Schoenberger2016CVPR}
Johannes~Lutz Schönberger and Jan-Michael Frahm.
\newblock Structure-from-motion revisited.
\newblock In \emph{CVPR}, 2016.

\bibitem[Shao et~al.(2025)Shao, Yang, Zhou, Zhang, Shen, Guizilini, Wang, Poggi, and Liao]{shao2024learning}
Jiahao Shao, Yuanbo Yang, Hongyu Zhou, Youmin Zhang, Yujun Shen, Vitor Guizilini, Yue Wang, Matteo Poggi, and Yiyi Liao.
\newblock Learning temporally consistent video depth from video diffusion priors.
\newblock In \emph{CVPR}, 2025.

\bibitem[Shi et~al.(2023)Shi, Chen, Zhang, Liu, Xu, Wei, Chen, Zeng, and Su]{Zero123++}
Ruoxi Shi, Hansheng Chen, Zhuoyang Zhang, Minghua Liu, Chao Xu, Xinyue Wei, Linghao Chen, Chong Zeng, and Hao Su.
\newblock Zero123++: a single image to consistent multi-view diffusion base model.
\newblock \emph{arXiv preprint arXiv:2310.15110}, 2023.

\bibitem[Shi et~al.(2024)Shi, Wang, Ye, Long, Li, and Yang]{Mvdream}
Yichun Shi, Peng Wang, Jianglong Ye, Mai Long, Kejie Li, and Xiao Yang.
\newblock Mvdream: Multi-view diffusion for 3d generation.
\newblock In \emph{ICLR}, 2024.

\bibitem[Snavely et~al.(2006)Snavely, Seitz, and Szeliski]{snavely2006photo}
Noah Snavely, Steven~M Seitz, and Richard Szeliski.
\newblock Photo tourism: exploring photo collections in 3d.
\newblock In \emph{SIGGRAPH}. 2006.

\bibitem[Song et~al.(2025{\natexlab{a}})Song, Li, Yang, Xu, Wei, Liu, Feng, Lin, and Zhang]{Puppeteer}
Chaoyue Song, Xiu Li, Fan Yang, Zhongcong Xu, Jiacheng Wei, Fayao Liu, Jiashi Feng, Guosheng Lin, and Jianfeng Zhang.
\newblock Puppeteer: Rig and animate your 3d models.
\newblock \emph{NeurIPS}, 2025{\natexlab{a}}.

\bibitem[Song et~al.(2025{\natexlab{b}})Song, Zhang, Li, Yang, Chen, Xu, Liew, Guo, Liu, Feng, et~al.]{Magicarticulate}
Chaoyue Song, Jianfeng Zhang, Xiu Li, Fan Yang, Yiwen Chen, Zhongcong Xu, Jun~Hao Liew, Xiaoyang Guo, Fayao Liu, Jiashi Feng, et~al.
\newblock Magicarticulate: Make your 3d models articulation-ready.
\newblock In \emph{CVPR}, 2025{\natexlab{b}}.

\bibitem[Sun et~al.(2024)Sun, Litvak, Zhang, Li, Wu, and Wu]{ponymation}
Keqiang Sun, Dor Litvak, Yunzhi Zhang, Hongsheng Li, Jiajun Wu, and Shangzhe Wu.
\newblock Ponymation: Learning articulated 3d animal motions from unlabeled online videos.
\newblock In \emph{ECCV}, 2024.

\bibitem[Tang et~al.(2024{\natexlab{a}})Tang, Chen, Chen, Wang, Zeng, and Liu]{Lgm}
Jiaxiang Tang, Zhaoxi Chen, Xiaokang Chen, Tengfei Wang, Gang Zeng, and Ziwei Liu.
\newblock Lgm: Large multi-view gaussian model for high-resolution 3d content creation.
\newblock In \emph{ECCV}, 2024{\natexlab{a}}.

\bibitem[Tang et~al.(2024{\natexlab{b}})Tang, Ren, Zhou, Liu, and Zeng]{Dreamgaussian}
Jiaxiang Tang, Jiawei Ren, Hang Zhou, Ziwei Liu, and Gang Zeng.
\newblock Dreamgaussian: Generative gaussian splatting for efficient 3d content creation.
\newblock In \emph{ICLR}, 2024{\natexlab{b}}.

\bibitem[Teed and Deng(2021)]{Droid}
Zachary Teed and Jia Deng.
\newblock Droid-slam: Deep visual slam for monocular, stereo, and rgb-d cameras.
\newblock \emph{NeurIPS}, 2021.

\bibitem[Unterthiner et~al.()Unterthiner, Van~Steenkiste, Kurach, Marinier, Michalski, and Gelly]{FVD}
Thomas Unterthiner, Sjoerd Van~Steenkiste, Karol Kurach, Rapha{\"e}l Marinier, Marcin Michalski, and Sylvain Gelly.
\newblock Fvd: A new metric for video generation.

\bibitem[Wang et~al.(2023{\natexlab{a}})Wang, Du, Li, Yeh, and Shakhnarovich]{sjc}
Haochen Wang, Xiaodan Du, Jiahao Li, Raymond~A Yeh, and Greg Shakhnarovich.
\newblock Score jacobian chaining: Lifting pretrained 2d diffusion models for 3d generation.
\newblock In \emph{CVPR}, 2023{\natexlab{a}}.

\bibitem[Wang et~al.(2025{\natexlab{a}})Wang, Chen, Karaev, Vedaldi, Rupprecht, and Novotny]{VGGT}
Jianyuan Wang, Minghao Chen, Nikita Karaev, Andrea Vedaldi, Christian Rupprecht, and David Novotny.
\newblock Vggt: Visual geometry grounded transformer.
\newblock In \emph{CVPR}, 2025{\natexlab{a}}.

\bibitem[Wang et~al.(2025{\natexlab{b}})Wang, Ye, Gao, Zeng, Austin, Li, and Kanazawa]{som}
Qianqian Wang, Vickie Ye, Hang Gao, Weijia Zeng, Jake Austin, Zhengqi Li, and Angjoo Kanazawa.
\newblock Shape of motion: 4d reconstruction from a single video.
\newblock In \emph{ICCV}, 2025{\natexlab{b}}.

\bibitem[Wang et~al.(2025{\natexlab{c}})Wang, Zhang, Holynski, Efros, and Kanazawa]{cut3r}
Qianqian Wang, Yifei Zhang, Aleksander Holynski, Alexei~A. Efros, and Angjoo Kanazawa.
\newblock Continuous 3d perception model with persistent state.
\newblock In \emph{CVPR}, 2025{\natexlab{c}}.

\bibitem[Wang et~al.(2024{\natexlab{a}})Wang, Leroy, Cabon, Chidlovskii, and Revaud]{dust3r}
Shuzhe Wang, Vincent Leroy, Yohann Cabon, Boris Chidlovskii, and Jerome Revaud.
\newblock {DUSt3R:} geometric 3d vision made easy.
\newblock In \emph{CVPR}, 2024{\natexlab{a}}.

\bibitem[Wang et~al.(2024{\natexlab{b}})Wang, Lipson, and Deng]{RAFT}
Yihan Wang, Lahav Lipson, and Jia Deng.
\newblock Sea-raft: Simple, efficient, accurate raft for optical flow.
\newblock In \emph{ECCV}, 2024{\natexlab{b}}.

\bibitem[Wang et~al.(2023{\natexlab{b}})Wang, Lu, Wang, Bao, Li, Su, and Zhu]{Prolificdreamer}
Zhengyi Wang, Cheng Lu, Yikai Wang, Fan Bao, Chongxuan Li, Hang Su, and Jun Zhu.
\newblock Prolificdreamer: High-fidelity and diverse text-to-3d generation with variational score distillation.
\newblock \emph{NeurIPS}, 2023{\natexlab{b}}.

\bibitem[Wu et~al.(2025{\natexlab{a}})Wu, Fang, Yang, Li, Yi, Lu, Zhou, Cen, Xie, Zhang, et~al.]{UniLat3D}
Guanjun Wu, Jiemin Fang, Chen Yang, Sikuang Li, Taoran Yi, Jia Lu, Zanwei Zhou, Jiazhong Cen, Lingxi Xie, Xiaopeng Zhang, et~al.
\newblock Unilat3d: Geometry-appearance unified latents for single-stage 3d generation.
\newblock \emph{arXiv preprint arXiv:2509.25079}, 2025{\natexlab{a}}.

\bibitem[Wu et~al.(2025{\natexlab{b}})Wu, Gao, Poole, Trevithick, Zheng, Barron, and Holynski]{Cat4d}
Rundi Wu, Ruiqi Gao, Ben Poole, Alex Trevithick, Changxi Zheng, Jonathan~T Barron, and Aleksander Holynski.
\newblock Cat4d: Create anything in 4d with multi-view video diffusion models.
\newblock In \emph{CVPR}, 2025{\natexlab{b}}.

\bibitem[Wu et~al.(2023)Wu, Li, Jakab, Rupprecht, and Vedaldi]{Magicpony}
Shangzhe Wu, Ruining Li, Tomas Jakab, Christian Rupprecht, and Andrea Vedaldi.
\newblock Magicpony: Learning articulated 3d animals in the wild.
\newblock In \emph{CVPR}, 2023.

\bibitem[Wu et~al.(2025{\natexlab{c}})Wu, Lin, Zhang, Zeng, Yang, Bao, Qian, Zhu, Cao, Torr, et~al.]{Direct3d-s2}
Shuang Wu, Youtian Lin, Feihu Zhang, Yifei Zeng, Yikang Yang, Yajie Bao, Jiachen Qian, Siyu Zhu, Xun Cao, Philip Torr, et~al.
\newblock Direct3d-s2: Gigascale 3d generation made easy with spatial sparse attention.
\newblock \emph{arXiv preprint arXiv:2505.17412}, 2025{\natexlab{c}}.

\bibitem[Wu et~al.(2025{\natexlab{d}})Wu, Zheng, Zhou, and Lu]{Point3R}
Yuqi Wu, Wenzhao Zheng, Jie Zhou, and Jiwen Lu.
\newblock Point3r: Streaming 3d reconstruction with explicit spatial pointer memory.
\newblock \emph{arXiv preprint arXiv:2507.02863}, 2025{\natexlab{d}}.

\bibitem[Wu et~al.(2025{\natexlab{e}})Wu, Yu, Wang, and Bai]{AnimateAnyMesh}
Zijie Wu, Chaohui Yu, Fan Wang, and Xiang Bai.
\newblock Animateanymesh: A feed-forward 4d foundation model for text-driven universal mesh animation.
\newblock In \emph{ICCV}, 2025{\natexlab{e}}.

\bibitem[Xiang et~al.(2025)Xiang, Lv, Xu, Deng, Wang, Zhang, Chen, Tong, and Yang]{TRELLIS}
Jianfeng Xiang, Zelong Lv, Sicheng Xu, Yu Deng, Ruicheng Wang, Bowen Zhang, Dong Chen, Xin Tong, and Jiaolong Yang.
\newblock Structured 3d latents for scalable and versatile 3d generation.
\newblock In \emph{CVPR}, 2025.

\bibitem[Xiao et~al.(2025)Xiao, Wang, Xue, Karaev, Makarov, Kang, Zhu, Bao, Shen, and Zhou]{Spatialtrackerv2}
Yuxi Xiao, Jianyuan Wang, Nan Xue, Nikita Karaev, Yuri Makarov, Bingyi Kang, Xing Zhu, Hujun Bao, Yujun Shen, and Xiaowei Zhou.
\newblock Spatialtrackerv2: 3d point tracking made easy.
\newblock In \emph{ICCV}, 2025.

\bibitem[Xie et~al.(2025)Xie, Yao, Voleti, Jiang, and Jampani]{Sv4d}
Yiming Xie, Chun-Han Yao, Vikram Voleti, Huaizu Jiang, and Varun Jampani.
\newblock Sv4d: Dynamic 3d content generation with multi-frame and multi-view consistency.
\newblock In \emph{ICLR}, 2025.

\bibitem[Xu et~al.(2023)Xu, Wang, Cheng, Cao, Shan, Qie, and Gao]{Dream3d}
Jiale Xu, Xintao Wang, Weihao Cheng, Yan-Pei Cao, Ying Shan, Xiaohu Qie, and Shenghua Gao.
\newblock Dream3d: Zero-shot text-to-3d synthesis using 3d shape prior and text-to-image diffusion models.
\newblock In \emph{CVPR}, 2023.

\bibitem[Xu et~al.(2024)Xu, Cheng, Gao, Wang, Gao, and Shan]{Instantmesh}
Jiale Xu, Weihao Cheng, Yiming Gao, Xintao Wang, Shenghua Gao, and Ying Shan.
\newblock Instantmesh: Efficient 3d mesh generation from a single image with sparse-view large reconstruction models.
\newblock \emph{arXiv preprint arXiv:2404.07191}, 2024.

\bibitem[Xu et~al.(2020)Xu, Zhou, Kalogerakis, Landreth, and Singh]{RigNet}
Zhan Xu, Yang Zhou, Evangelos Kalogerakis, Chris Landreth, and Karan Singh.
\newblock Rignet: Neural rigging for articulated characters.
\newblock \emph{ACM Trans. on Graphics}, 2020.

\bibitem[Yang et~al.(2024)Yang, Gao, Zhou, Jiao, Zhang, and Jin]{Deformable3dgaussians}
Ziyi Yang, Xinyu Gao, Wen Zhou, Shaohui Jiao, Yuqing Zhang, and Xiaogang Jin.
\newblock Deformable 3d gaussians for high-fidelity monocular dynamic scene reconstruction.
\newblock In \emph{CVPR}, 2024.

\bibitem[Yang et~al.(2025)Yang, Pan, Gu, and Zhang]{yang2024diffusion}
Zeyu Yang, Zijie Pan, Chun Gu, and Li Zhang.
\newblock Diffusion2: Dynamic 3d content generation via score composition of video and multi-view diffusion models.
\newblock In \emph{ICLR}, 2025.

\bibitem[Yao et~al.(2025)Yao, Xie, Voleti, Jiang, and Jampani]{Sv4d2}
Chun-Han Yao, Yiming Xie, Vikram Voleti, Huaizu Jiang, and Varun Jampani.
\newblock Sv4d 2.0: Enhancing spatio-temporal consistency in multi-view video diffusion for high-quality 4d generation.
\newblock In \emph{ICCV}, 2025.

\bibitem[Yenphraphai et~al.(2025)Yenphraphai, Mirzaei, Chen, Zou, Tulyakov, Yeh, Wonka, and Wang]{ShapeGen4D}
Jiraphon Yenphraphai, Ashkan Mirzaei, Jianqi Chen, Jiaxu Zou, Sergey Tulyakov, Raymond~A Yeh, Peter Wonka, and Chaoyang Wang.
\newblock Shapegen4d: Towards high quality 4d shape generation from videos.
\newblock \emph{arXiv preprint arXiv:2510.06208}, 2025.

\bibitem[Yi et~al.(2024)Yi, Fang, Wang, Wu, Xie, Zhang, Liu, Tian, and Wang]{Gaussiandreamer}
Taoran Yi, Jiemin Fang, Junjie Wang, Guanjun Wu, Lingxi Xie, Xiaopeng Zhang, Wenyu Liu, Qi Tian, and Xinggang Wang.
\newblock Gaussiandreamer: Fast generation from text to 3d gaussians by bridging 2d and 3d diffusion models.
\newblock In \emph{CVPR}, 2024.

\bibitem[Yu et~al.(2018)Yu, Liu, Liu, Xie, Yang, Wei, and Qiao]{DS-SLAM}
Chao Yu, Zuxin Liu, Xin{-}Jun Liu, Fugui Xie, Yi Yang, Qi Wei, and Fei Qiao.
\newblock {DS-SLAM:} {A} semantic visual {SLAM} towards dynamic environments.
\newblock In \emph{IROS}, 2018.

\bibitem[Zhang et~al.(2023)Zhang, Tang, Niessner, and Wonka]{3dshape2vecset}
Biao Zhang, Jiapeng Tang, Matthias Niessner, and Peter Wonka.
\newblock 3dshape2vecset: A 3d shape representation for neural fields and generative diffusion models.
\newblock \emph{ACM Trans. on Graphics}, 2023.

\bibitem[Zhang et~al.(2025{\natexlab{a}})Zhang, Xu, Wang, Yang, Zhao, Chen, and Guo]{gvfd}
Bowen Zhang, Sicheng Xu, Chuxin Wang, Jiaolong Yang, Feng Zhao, Dong Chen, and Baining Guo.
\newblock Gaussian variation field diffusion for high-fidelity video-to-4d synthesis.
\newblock In \emph{ICCV}, 2025{\natexlab{a}}.

\bibitem[Zhang et~al.(2024{\natexlab{a}})Zhang, Chen, Wang, Liu, Wang, and Qiao]{4diffusion}
Haiyu Zhang, Xinyuan Chen, Yaohui Wang, Xihui Liu, Yunhong Wang, and Yu Qiao.
\newblock 4diffusion: Multi-view video diffusion model for 4d generation.
\newblock \emph{NeurIPS}, 2024{\natexlab{a}}.

\bibitem[Zhang et~al.(2025{\natexlab{b}})Zhang, Herrmann, Hur, Jampani, Darrell, Cole, Sun, and Yang]{monst3r}
Junyi Zhang, Charles Herrmann, Junhwa Hur, Varun Jampani, Trevor Darrell, Forrester Cole, Deqing Sun, and Ming-Hsuan Yang.
\newblock {MonST3R:} a simple approach for estimating geometry in the presence of motion.
\newblock In \emph{ICLR}, 2025{\natexlab{b}}.

\bibitem[Zhang et~al.(2025{\natexlab{c}})Zhang, Pu, Guo, Cao, and Hu]{UniRig}
Jia-Peng Zhang, Cheng-Feng Pu, Meng-Hao Guo, Yan-Pei Cao, and Shi-Min Hu.
\newblock One model to rig them all: Diverse skeleton rigging with unirig.
\newblock \emph{ACM Trans. on Graphics}, 2025{\natexlab{c}}.

\bibitem[Zhang et~al.(2024{\natexlab{b}})Zhang, Wang, Zhang, Qiu, Pang, Jiang, Yang, Xu, and Yu]{Clay}
Longwen Zhang, Ziyu Wang, Qixuan Zhang, Qiwei Qiu, Anqi Pang, Haoran Jiang, Wei Yang, Lan Xu, and Jingyi Yu.
\newblock Clay: A controllable large-scale generative model for creating high-quality 3d assets.
\newblock \emph{ACM Trans. on Graphics}, 2024{\natexlab{b}}.

\bibitem[Zhang et~al.(2018)Zhang, Isola, Efros, Shechtman, and Wang]{LPIPS}
Richard Zhang, Phillip Isola, Alexei~A Efros, Eli Shechtman, and Oliver Wang.
\newblock The unreasonable effectiveness of deep features as a perceptual metric.
\newblock In \emph{CVPR}, 2018.

\bibitem[Zhang et~al.(2022)Zhang, Cole, Li, Snavely, Rubinstein, and Freeman]{CasualSAM}
Zhoutong Zhang, Forrester Cole, Zhengqi Li, Noah Snavely, Michael Rubinstein, and William~T. Freeman.
\newblock Structure and motion from casual videos.
\newblock In \emph{ECCV}, 2022.

\bibitem[Zhao et~al.(2025{\natexlab{a}})Zhao, Nag, Wang, Vora, Ji, Chun, Mahdavi-Amiri, and Zhang]{zhao2025advances}
Mingrui Zhao, Sauradip Nag, Kai Wang, Aditya Vora, Guangda Ji, Peter Chun, Ali Mahdavi-Amiri, and Hao Zhang.
\newblock Advances in 4d representation: Geometry, motion, and interaction.
\newblock \emph{arXiv preprint arXiv:2510.19255}, 2025{\natexlab{a}}.

\bibitem[Zhao et~al.(2022)Zhao, Liu, Guo, Wang, and Liu]{zhao2022particlesfm}
Wang Zhao, Shaohui Liu, Hengkai Guo, Wenping Wang, and Yong-Jin Liu.
\newblock Particlesfm: Exploiting dense point trajectories for localizing moving cameras in the wild.
\newblock In \emph{ECCV}, 2022.

\bibitem[Zhao et~al.(2023)Zhao, Yan, Xie, Hong, Li, and Lee]{Animate124}
Yuyang Zhao, Zhiwen Yan, Enze Xie, Lanqing Hong, Zhenguo Li, and Gim~Hee Lee.
\newblock Animate124: Animating one image to 4d dynamic scene.
\newblock \emph{arXiv preprint arXiv:2311.14603}, 2023.

\bibitem[Zhao et~al.(2025{\natexlab{b}})Zhao, Lai, Lin, Zhao, Liu, Yang, Feng, Yang, Zhang, Yang, et~al.]{Hunyuan3d_2}
Zibo Zhao, Zeqiang Lai, Qingxiang Lin, Yunfei Zhao, Haolin Liu, Shuhui Yang, Yifei Feng, Mingxin Yang, Sheng Zhang, Xianghui Yang, et~al.
\newblock Hunyuan3d 2.0: Scaling diffusion models for high resolution textured 3d assets generation.
\newblock \emph{arXiv preprint arXiv:2501.12202}, 2025{\natexlab{b}}.

\bibitem[Zheng et~al.(2024{\natexlab{a}})Zheng, Gao, Fan, Liu, Laaksonen, Ouyang, and Sebe]{biref}
Peng Zheng, Dehong Gao, Deng-Ping Fan, Li Liu, Jorma Laaksonen, Wanli Ouyang, and Nicu Sebe.
\newblock Bilateral reference for high-resolution dichotomous image segmentation.
\newblock \emph{CAAI AIR}, 3:\penalty0 1--12, 2024{\natexlab{a}}.

\bibitem[Zheng et~al.(2024{\natexlab{b}})Zheng, Li, Nagano, Liu, Hilliges, and De~Mello]{zheng2024unified}
Yufeng Zheng, Xueting Li, Koki Nagano, Sifei Liu, Otmar Hilliges, and Shalini De~Mello.
\newblock A unified approach for text-and image-guided 4d scene generation.
\newblock In \emph{CVPR}, 2024{\natexlab{b}}.

\bibitem[Zhuo et~al.(2025)Zhuo, Zheng, Guo, Wu, Zhou, and Lu]{StreamVGGT}
Dong Zhuo, Wenzhao Zheng, Jiahe Guo, Yuqi Wu, Jie Zhou, and Jiwen Lu.
\newblock Streaming 4d visual geometry transformer.
\newblock \emph{arXiv preprint arXiv:2507.11539}, 2025.

\end{thebibliography}
}

\clearpage
\setcounter{page}{1}
\maketitlesupplementary
\appendix

\section{Implementation Details}
\label{sec:Implementation}
\subsection{Training and Inference Details}
\paragraph{Network Architecture.} Our network architecture consists of two primary components: a shape encoder and a motion latent block. For the shape encoder, we adopt the architecture from 3DShape2VecSet~\cite{3dshape2vecset}, which provides strong representational capacity and high compression efficiency. Specifically, we employ a point cloud encoder to extract features from the canonical shape representation. The encoder takes $N = 4096$ sampled points as input. First, the point coordinates are projected through a Fourier feature embedding layer to obtain position embeddings. These embeddings are then concatenated with surface normals and color attributes, and subsequently projected to the model dimension $d=768$ via a linear layer. To aggregate point features into a compact representation, we employ a QK-Norm cross-attention block with 64 learnable query tokens. The resulting tokens are further processed through 4 transformer layers with self-attention mechanisms to capture inter-point relationships, ultimately producing a semantically rich canonical shape representation $\bZ \in \mathbb{R}^{64 \times 768}$.

For video encoding, we adopt a frozen DINOv2-ViT-B/14 model~\cite{dinov2} to extract spatial features from each frame. Input videos are resized to $224 \times 224$ resolution and processed frame-by-frame through the DINOv2 encoder, which extracts $256$ patch tokens per frame with dimension 768. For a video with $T$ frames, we obtain image tokens $\bV \in \mathbb{R}^{T \times 256 \times 768}$. These tokens are combined with fixed positional embeddings generated using sinusoidal encoding over both temporal and spatial dimensions. To further aggregate spatio-temporal information from the video, we process these embeddings through Motion 3-to-4 blocks. Following the design of VGGT~\cite{VGGT}, we adopt an alternating global-frame attention architecture consisting of 16 layers (8 global and 8 frame). Both global and frame attention layers use QK-normalization. This alternating design effectively captures both spatial and temporal dependencies while maintaining computational efficiency.

To predict per-point motion trajectories, we extract 64 motion tokens from the processed video sequence. For each point in $M = 4096$ output point clouds, we construct a query using its position, normal, and color through the same embedding layer used in the shape encoder. We then apply a QK-Norm cross-attention layer where the per-point queries attend to the extracted motion tokens from the corresponding frame. This cross-attention mechanism produces per-point features $\bF \in \mathbb{R}^{M \times 768}$ that encode motion information. Finally, a two-layer MLP with GELU activation decodes these features into motion trajectories $\bX_t \in \mathbb{R}^{M \times 3}$ for time t. During inference, when we need to animate mesh vertices, we process them in chunks of 4096 to maintain memory efficiency. For temporal processing, we adopt a chunk size of 256 frames. When dealing with videos exceeding 256 frames, we use a sliding window approach with a stride of 255 frames, where each window consists of the first frame concatenated with 255 subsequent frames to maintain temporal consistency.
\vspace{-5mm}
\paragraph{Training Configuration.} We employ the AdamW optimizer with learning rate $\eta = 4 \times 10^{-4}$, $\beta_1=0.9$, $\beta_2=0.95$, and weight decay $0.05$. The learning rate follows a cosine annealing schedule with 1000 warm-up steps. We train for 60,000 parameter update steps with gradient clipping at norm 1.0. To reduce memory consumption and accelerate training, we apply gradient checkpointing, FlashAttention-v2 in the xFormers library, and a mixed precision strategy with BF16.
\vspace{-5mm}
\paragraph{Training Data and Strategy.} For training data, we use videos at $256\times256$ resolution with black backgrounds. The point clouds are uniformly sampled on the mesh surface, and crucially, we sample points at consistent barycentric coordinates within each face across all frames. This ensures temporal correspondence between points across different frames, enabling us to track each point's trajectory and supervise the model using MSE loss.

For the training strategy, we train on 12-frame sequences and apply temporal data augmentation. Specifically, we randomly select a starting frame and then sample 12 consecutive frames with stride intervals of 1, 2, or 4 frames. This augmentation strategy enables the model to handle different initial poses and learn to predict larger motion displacements across varying temporal scales.
\vspace{-5mm}
\paragraph{Inference with video.} For cases where only video input is provided without any mesh, we leverage Hunyuan 2.0~\cite{Hunyuan3d_2} to generate a mesh based on the first frame. It is worth noting that the directly generated mesh is non-watertight and includes texture. To address the watertight issue, we apply a vertex mapping technique to convert it into a watertight mesh while preserving the original texture, which is essential for subsequent video-driven animation.
For cases where a mesh is already provided, we directly drive the mesh using generated or existing videos.
\vspace{-2mm}
\subsection{Evaluation Details}
We utilize the official release code for evaluation. For our held-out dataset, we use the front view as the input view and the remaining three orthogonal views for evaluation. We exclude the front view from evaluation because including it would be unfair to generation-based methods, since L4GM~\cite{L4gm} can perform lossless reconstruction of the input view (i.e., the front view). 

In the \tabref{tab:mesh}, "OOM" refers to "out of memory." For the GVFD~\cite{gvfd}, the official released code does not provide scripts for processing long sequences; it only includes scripts for single-video inference. Consequently, when the sequence length exceeds 128 frames, our machine encounters the out-of-memory (OOM) problem, preventing us from obtaining quantitative results for this method.
 \vspace{-2mm}
\subsection{Ablation Study}
 \vspace{-1mm}
We conduct an ablation study to explore different model architecture choices. Due to limited training resources, we train each model variant for 30,000 steps under identical settings. To evaluate the trajectory prediction capability of each variant, we report the mean squared error (MSE) of point trajectories over time on the held-out dataset in \tabref{tab:abl}, which provides a more direct measure of the model's ability to track accurate motion trajectories.

\begin{table}[h]
  \centering
  \resizebox{0.8\columnwidth}{!}{
    \begin{tabular}{ccc|c}
      \toprule
      $Frame~Attn$ & $Global~Attn$ & $Ref~Token$ & Rec MSE $\downarrow$ \\
      \midrule
      \xmark & \cmark & \cmark & 0.0055 \\          %
      \cmark & \xmark & \cmark & 0.0033 \\          %
      \cmark & \cmark & \xmark & 0.0021 \\          %
      \cmark & \cmark & \cmark & \textbf{0.0018} \\ %
      \bottomrule
    \end{tabular}
  }
  \vspace{-.1in}
  \caption{Ablation studies of the modules of Motion 3-to-4. Rec MSE denotes the squared error averaged across time steps within the $[-0.5,0.5]$ bounding box.}
  \label{tab:abl}
  \vspace{-.1in}
\end{table}
 \vspace{-3mm}
\section{More Results}
\label{sec:results}

We provide several categories of additional visualization results below. We also include a local video webpage in the supplementary materials to better present the results.
 \vspace{-2mm}
\subsection{More Results from Synthesis Video}
\begin{figure}[h]
  \centering
  \includegraphics[width=\linewidth]{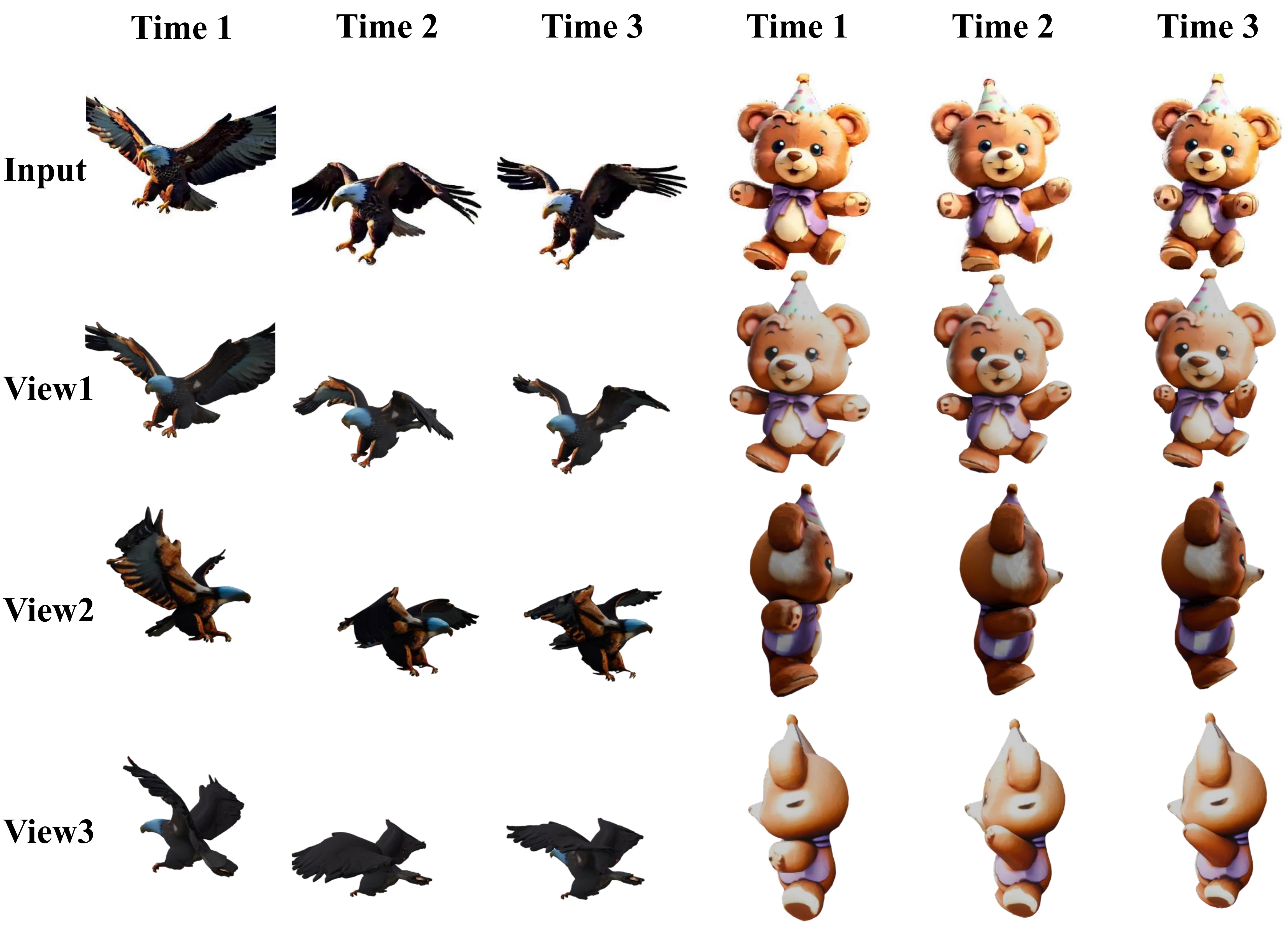}
  \vspace{-5mm}
  \caption{Additional visual results from synthesis video.}
  \label{fig:sync}
\end{figure}

As illustrated in \figref{fig:sync}, our model demonstrates strong generalization capability across diverse test cases from synthetic videos, achieving high-quality results that showcase its ability to handle various object types and motion patterns.
 \vspace{-2mm}
\subsection{More Results from Real-World Video}
\begin{figure}[h]
  \centering
  \includegraphics[width=\linewidth]{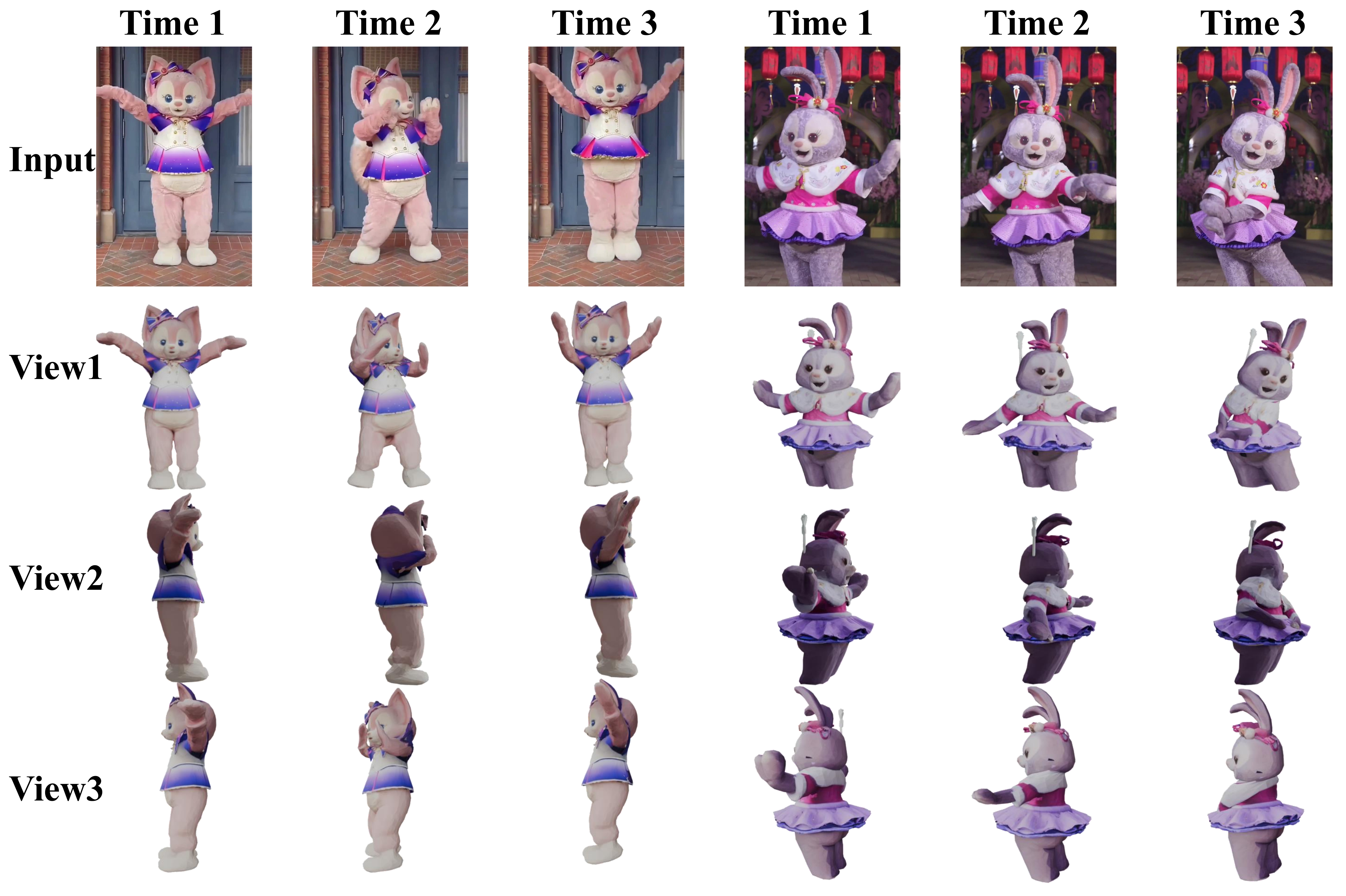}
  \vspace{-5mm}
  \caption{Additional visual results from real-world video.}
  \vspace{-5mm}
  \label{fig:real}
\end{figure}

As illustrated in \figref{fig:real}, despite being trained exclusively on synthetic data, our model generalizes well to real-world videos. This demonstrates the model's robustness and its ability to bridge the synthetic-to-real domain gap, highlighting the effectiveness of our method.
 \vspace{-2mm}
\subsection{Results of Existing 3D Model}
\begin{figure}[h]
  \centering
  \includegraphics[width=\linewidth]{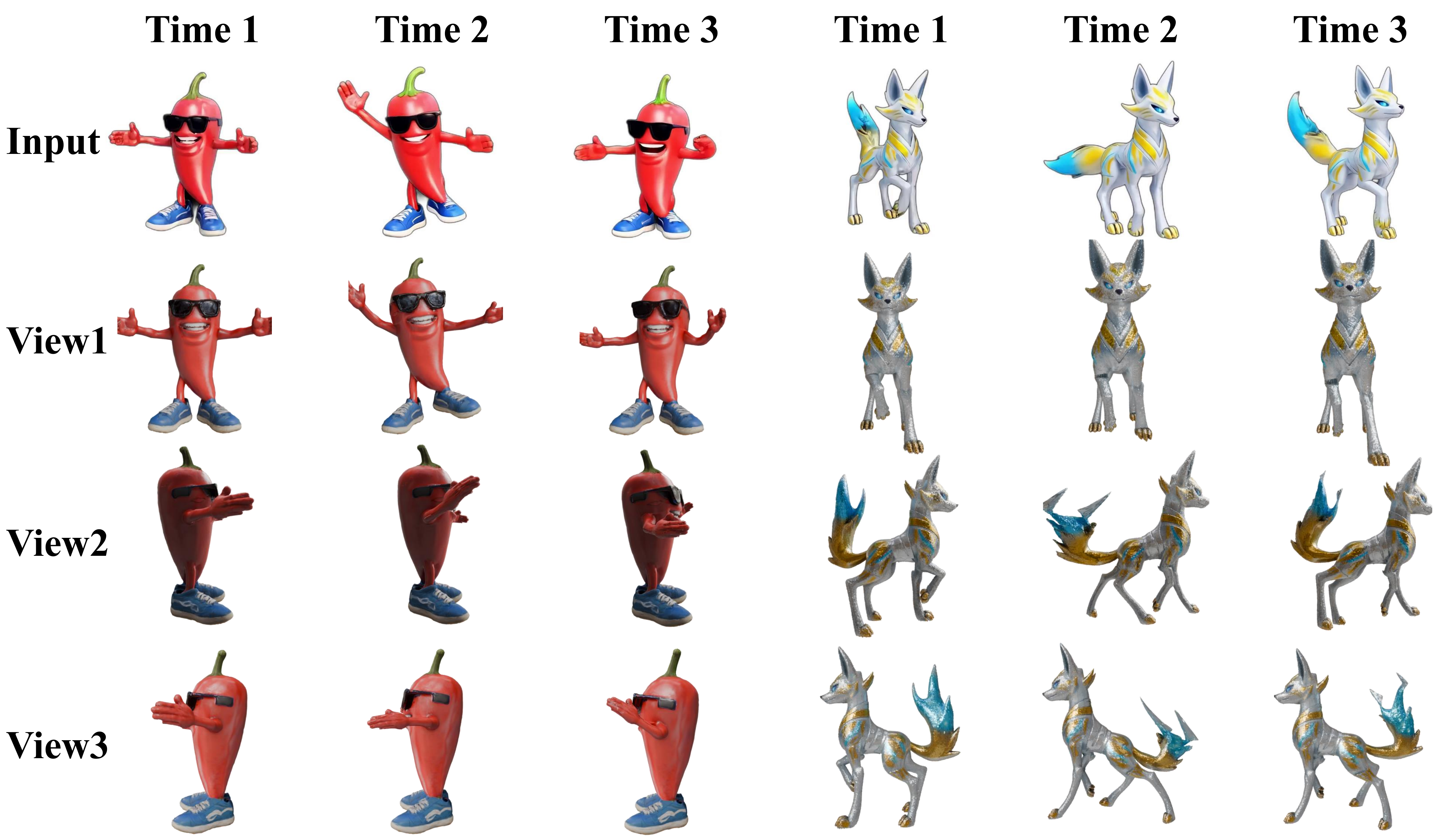}
  \vspace{-5mm}
  \caption{Results of existing 3D model condition on generated video.}
  \label{fig:3da}
  \vspace{-5mm}
\end{figure}

As illustrated in \figref{fig:3da}, our model has strong practical value as it can extend existing static 3D assets to dynamic 4D content. This capability enables flexible application scenarios, such as animating existing 3D models from various sources, thereby significantly broadening the potential use cases of our method.

\begin{figure*}[t]
  \centering
  \includegraphics[width=1.0\linewidth]{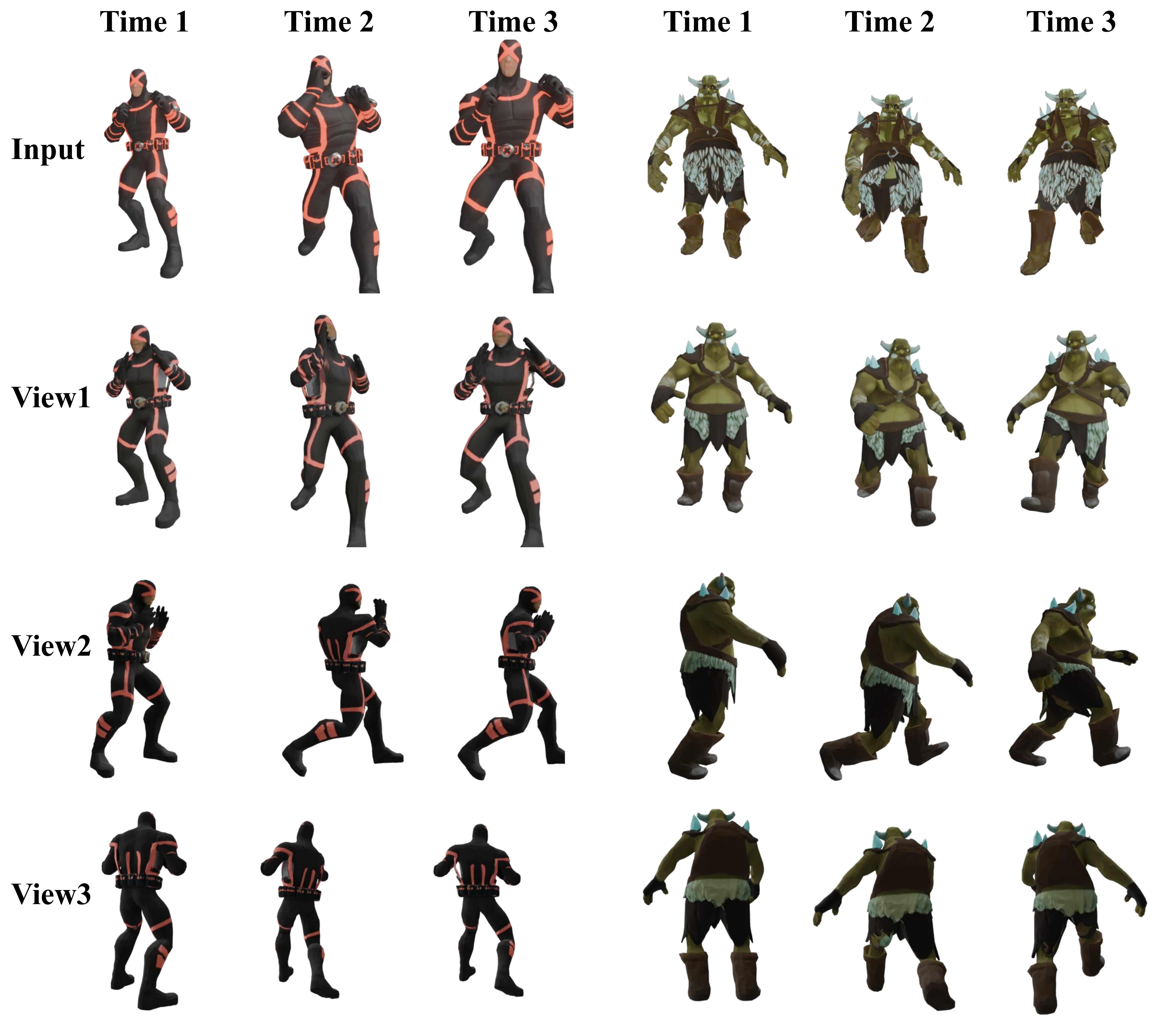}
  \caption{More results from the held-out objaverse dataset.}
  \label{fig:objmore}
  \vspace{-3mm}
\end{figure*}
\subsection{Visual Comparison on Consist4D}
\begin{figure*}[t]
  \centering
  \includegraphics[width=1.0\linewidth]{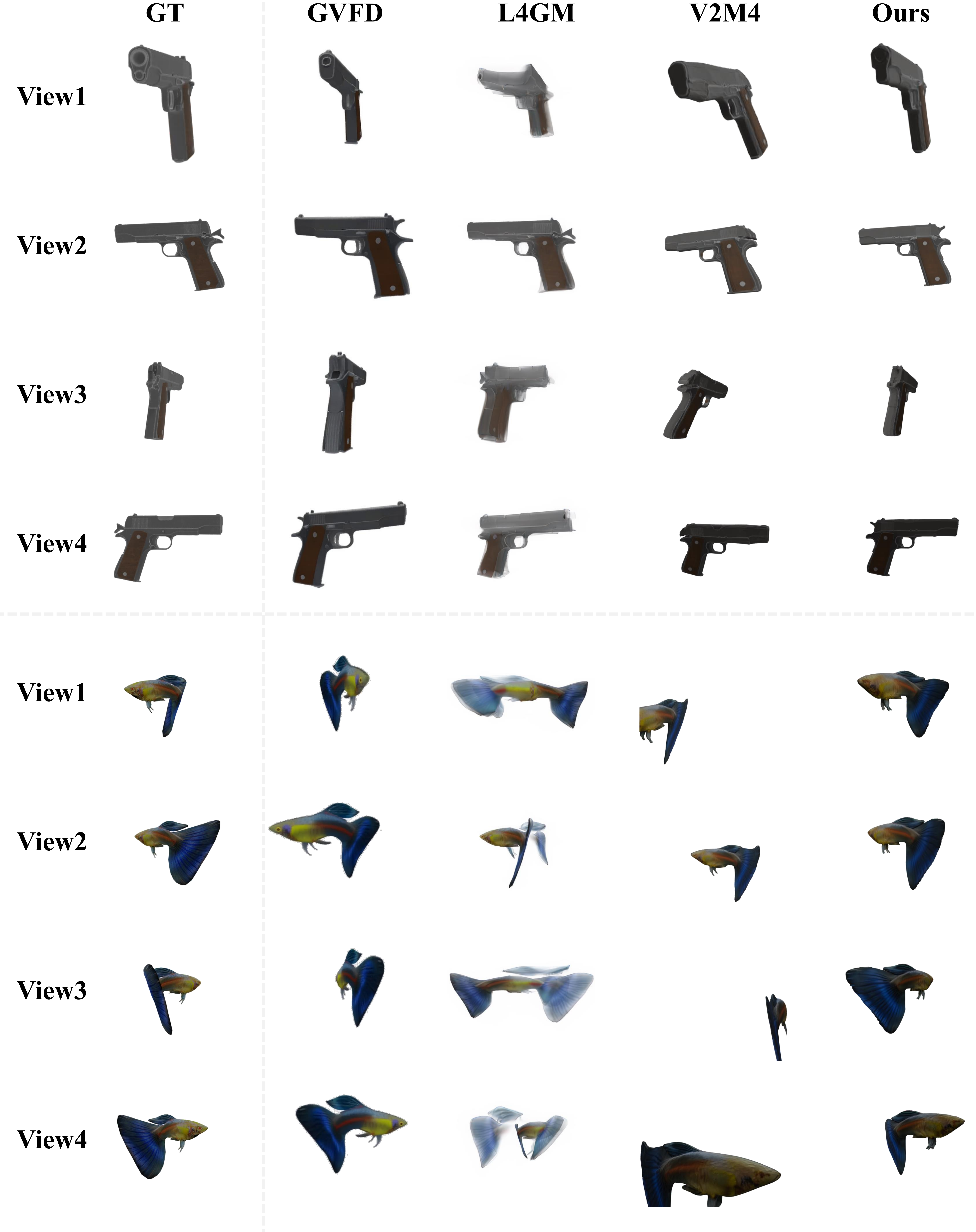}
  \caption{Visual comparison with SOTA methods on Consist4D.}
  \label{fig:consist4dv}
\end{figure*}

The Consist4D dataset~\cite{Consistent4d} provides input views from various angles that differ from the frontal views in our held-out dataset, which better demonstrates the model's robustness to different viewpoints. This is especially relevant for method L4GM, which is designed to work well under orthogonal views. When the input view is not orthogonal, L4GM's multi-view diffusion module fails to generate consistent results across views, leading to severe ghosting artifacts as shown in Figure~\ref{fig:consist4dv}.

\subsection{More Results from the Held-Out Dataset}

As illustrated in \figref{fig:objmore}, we present additional results on our held-out dataset to demonstrate that our model achieves superior visual fidelity and temporal motion coherence. These results further validate the effectiveness of our approach in generating high-quality 4D content with both realistic appearance and consistent motion over time.

\end{document}